\documentclass{article}
\usepackage{microtype}
\usepackage{graphicx}
\usepackage{booktabs}
\usepackage{hyperref}

\usepackage[preprint]{icml2026}
\usepackage{amsmath,amssymb,amsfonts,amsthm}
\usepackage[capitalize,noabbrev]{cleveref}
\usepackage[table]{xcolor}

\icmltitlerunning{Concept vectors reflect only the symmetric part of the conceptual structure of mathematics}

\begin{document}

\twocolumn[
  \icmltitle{Concept vectors reflect only the symmetric part\\
    of the conceptual structure of mathematics}
  \icmlsetsymbol{equal}{*}
  \begin{icmlauthorlist}
    \icmlauthor{Silv\`ere Gangloff}{ostrava}
  \end{icmlauthorlist}
  \icmlaffiliation{ostrava}{University of Ostrava, Ostrava, Czech Republic}
  \icmlcorrespondingauthor{Silv\`ere Gangloff}{silvere.f.gangloff@gmail.com}
  \icmlkeywords{interpretability, concept vectors, steering, mathematics}
  \vskip 0.3in
]

\printAffiliationsAndNotice{}

\begin{abstract}
Macro-interpretability studies how language models represent
concepts through populations of activations, without seeking to
describe individual circuits. One of its foundational notions is the concept
vector, the mean difference between a model's activations on texts about a
concept and on other texts. Where concepts lack exact definitions, as
for emotions, concept vectors are validated by steering. Mathematics,
whose concepts are exactly defined and related, allows a finer
assessment. We find that steering validates mathematical concept
vectors only weakly: its effect on the text is confined to a small window of
intensities and to texts where the concept is already relevant, and,
unlike for emotions, these vectors are nearly orthogonal to the
direction that most strongly writes the name of the concept. We
therefore assess them against the structure of mathematics, through the
inclusion graph of its concepts. We find that concept vectors reflect the
distance between concepts in this graph, but not the direction in it,
although the model itself represents it. We argue that this limitation lies in the
choices that define concept vectors, and that conceptual structure provides
a yardstick for refining them.
\end{abstract}

\section{Introduction}
\label{sec:intro}

Interpretability of language models is pursued at two levels.
Micro-interpretability consists in reverse-engineering the specific
circuits or algorithms behind a certain capability. This is a powerful
lens when it succeeds. However, a particular circuit is a priori
specific to the trained model, and not to the capability alone. For
instance, \citet{zhong2023clock} show that modular addition, as a
capability, is implemented by different algorithms depending on the
training seed. Macro-interpretability instead represents concepts by population-level directions in the
model's activations, called \emph{concept vectors}, typically the mean
difference between activations on texts about a concept and on other
texts. It comprises studies on emotions
\citep{sofroniew2026emotion}, character traits \citep{chen2025persona},
refusal \citep{arditi2024refusal} and general alignment-relevant
behavior \citep{zou2023representation, turner2023steering,
rimsky2024steering}.
Macro-interpretability claims are coarser and more replicable than the
description of a circuit, at the cost of not explaining the
computational processes underlying the model's behavior. Neuroscience
has developed both levels of description extensively. At the level of
systems, neuroscience studies which regions or networks support which
functions, without resolving individual mechanisms.

In language models, interpretability work on mathematics is mostly
at the microscopic level \citep{nanda2023progress, zhong2023clock,
stolfo2023mechanistic, nikankin2025arithmetic, anthropic2025biology},
and more recently at the macroscopic level \citep{zhou2024fourier,
kantamneni2025trigonometry, wen2026shape, anthropic2026jlens}, in both
cases mainly on arithmetic. To our knowledge, macro-interpretability
has not been applied to abstract conceptual structures, such as mathematics, in which concepts have exact
definitions and are related in definite ways, and whether a fixed
object has a fixed property is also definite. On the other hand, in neuroscience,
mathematics has been studied at both levels, from single neurons tuned
to numbers \citep{nieder2002quantity} to the networks of brain areas
recruited by advanced mathematics \citep{amalric2016origins}.
The exactness of the structure makes it possible to question how
much of it concept vectors reflect, and thus to assess similar notions
more finely than is possible in domains without exact structure.

In domains without exact structure, such as those studied so far,
where no ground truth is available, methods that attach a direction of the model's
activations to an interpretable property of the text, such as concept
vectors, are validated by steering. Adding the direction to the activations during generation
is expected to make the outputs exhibit this property more strongly. We believe that this validation rests on an
intuition from human cognition: understanding a concept means being able to use it, so a
direction that makes the model express a concept is taken to be a direction along which the model represents
it. However, nothing requires these two domains
of directions to overlap significantly for either
of them to be meaningful. This distinction is familiar for
systems neuroscience, where showing that the brain uses a signal
requires intervening on it, beyond decoding it
\citep{dewit2016information,jazayeri2017navigating}, and for earlier
interpretability work (\cref{sec:related}).

We also call the concept vector the \emph{reading vector}, and we
distinguish one of the directions that write the concept, which we call
the \emph{writing vector}: the direction that most strongly makes the
model write the name of the concept.
Macro-interpretability takes the alignment of the two, observed for
emotions, as a sign that the concept vector is a faithful readout of
the concept. We study their relation in
mathematics. We find that steering validates mathematical
concept vectors only weakly: it changes the text only in a small window
of intensities, and only where the concept is already relevant. How
much a concept vector writes the name of its concept follows its cosine
with the writing vector of the name, which is several times smaller
for mathematical concepts than for emotions (\cref{sec:writing}).

We instead assess concept vectors by how much of the structure of mathematics
they preserve.
We consider the conceptual structure of mathematics through the
inclusion graph of definitions, whose vertices are mathematical
structures, with an edge from one structure to another when every
instance of the former is an instance of the latter. We find that concept
vectors reflect what we call the symmetric part of the structure of
this graph: how close two concepts are in it, which does not depend on
the order in which the two concepts are taken. They do not reflect its
asymmetric part, which depends on this order: the direction of its
edges, that is, which of two concepts includes the other
(\cref{sec:geometry}). Nor do they reflect a closely related
asymmetric relation, similar to the one that defines the graph, which
relates objects to the structures they are instances of
(\cref{sec:aboutness}). On the other hand, the model itself reflects this asymmetry. A
probe on its activations tells whether an object is an instance of a
structure. Furthermore, a variation of the definition of the concept
vector reflects it. This variation averages questions about objects, read
at the token where the object is named, and subtracts the mean over
questions about objects that are not instances of the structure
(\cref{sec:aboutness}). This implies that the limitation comes from the
choices of parameters in the definition of the concept vector. Contrary to the natural intuition that a macroscopic method, which
averages over populations of activations, cannot capture mathematical
concepts, concept vectors are thus relevant to
mathematical concepts: they capture part of their structure, even if
only part of it.

A second way to delimit what concept vectors describe is to compare
them with simple descriptions of the texts they are computed from, as
is standard in neuroscience \citep{kriegeskorte2008rsa,khaligh2014deep}
and in probing \citep{conneau2018cram}. We compute every result also
from classical statistics of the words of the texts
(\cref{sec:method}). Word statistics reflect the symmetric part of the
structure as well as concept vectors do, or better, and once the part
of the concept vectors that word statistics explain is removed,
nothing of this symmetric part remains (\cref{sec:geometry}). For the
asymmetric part, the comparison goes the other way: the probe that
captures it on the model's activations fails on word statistics
(\cref{sec:aboutness}). This could be expected, since word statistics
ignore the order of words, and thus cannot tell apart ``every $S_1$ is
an $S_2$'' from ``every $S_2$ is an $S_1$'', for two structures $S_1$
and $S_2$.

\textbf{Contributions.}
\begin{itemize}
  \item Evidence that steering is at best a weak validation of
    mathematical concept vectors, as they write their concept only in a
    narrow window of intensities (\cref{sec:writing}).
  \item A partial validation of concept vectors in mathematics, as they
    reflect part of its conceptual structure (\cref{sec:geometry}).
  \item Evidence that word statistics of the same texts already capture
    what concept vectors reflect of this structure (\cref{sec:geometry}).
  \item Evidence that the models we study represent more about
    mathematical concepts than what concept vectors capture, and that
    variants of concept vectors capture part of it
    (\cref{sec:aboutness}).
\end{itemize}

\section{Related work}
\label{sec:related}

\textbf{Concept vectors and steering.} To define concept vectors, we follow the
contrastive mean-difference recipe used for emotions
\citep{sofroniew2026emotion}, character traits \citep{chen2025persona}
and temporal discounting \citep{riossialer2026temporal}, in the
line of activation steering
\citep{turner2023steering,rimsky2024steering,zou2023representation,arditi2024refusal}. We center the vectors, which corresponds to mean-centring for
steering \citep{jorgensen2023meancentring} and to removing the common directions of anisotropic representations
\citep{mu2018allbutthetop,timkey2021rogue}. In addition to
steering, which validates directions in this line of work, we assess concept vectors by how much of the
inclusion graph of mathematical definitions they reflect, and compare
each result with the same measure on word statistics of the texts.

\textbf{Reading and writing.} That a direction which reads a
concept need not be one that writes it is known. Representation
engineering separates representation reading from representation
control \citep{zou2023representation}. In inference-time intervention,
the mass-mean direction steers better than the weights of the best
probe \citep{li2023iti}, and mass-mean probes are more causally
implicated than logistic-regression probes of similar accuracy
\citep{marks2024geometry}. On AxBench, difference-in-means
directions are the best concept detectors but steer less well than
prompting or finetuning \citep{wu2025axbench}.

\textbf{Concept geometry.}  \citet{park2024linear} argue
that concepts such as the language of a word are well represented
by directions. \citet{park2025geometry} extend
this to hierarchies of categories of words, and find that, for a
suitable inner product, the vector of a category is orthogonal to its
difference with the vector of any of its descendants. With their
estimator, this orthogonality does not hold for the concepts of the
inclusion graph (\cref{sec:inclusion}), which are named by several
tokens and defined by texts. \citet{elhage2022toy} show that a model
can represent a feature without devoting a direction to it, so a
concept vector can miss part of what the model represents about a
concept.

\textbf{Mathematics in language models.} Most interpretability
work on mathematics is micro-level and studies arithmetic.
Circuits have been found for modular addition
\citep{nanda2023progress,zhong2023clock} and for addition
\citep{anthropic2025biology}, and arithmetic is carried by specific
components \citep{stolfo2023mechanistic} and by a bag of heuristics in
individual neurons \citep{nikankin2025arithmetic}. The Jacobian lens
\citep{anthropic2026jlens} reads intermediate results. Closest in
spirit, \citet{venhoff2025understanding} build steering vectors for
reasoning behaviors such as backtracking and verification, and
\citet{wen2026shape} study the geometry of multi-operand addition.
None of this work studies mathematical concepts with the methods
of macro-interpretability.

\section{Setup and methods}
\label{sec:method}

\Cref{app:methods} details the material and methods that the
experiments share.

\paragraph{Concept bank.} We built a \emph{concept bank} of 201
mathematical concepts of six kinds: 101
\emph{structures} (group, ring, metric space, tree, \dots), 35
\emph{techniques} (induction, the pigeonhole principle, \dots), 24
\emph{constructions}, each obtained by applying one of four relations
(sub-object, morphism, quotient, product) to one of six structures
(for instance, the quotient ring), 18 \emph{instances} ($\mathbb{Z}$,
$S_3$, the complete graph $K_5$, \dots), each listing the structures
it belongs to, 14 properties and operations used in definitions
(associativity, inverses, \dots) and 9 theorems (\cref{app:bank}).

\paragraph{Inclusion graph.} We denote by $G$ the inclusion graph on
the structures of the bank. It has an edge from a structure to another
when every instance of the first is an instance of the second, and
this does not follow from other edges by composition. The \emph{ancestors}
of a structure are the structures reached from it by a directed path
in $G$. A structure is a \emph{descendant} of each of its ancestors,
and two structures are \emph{unrelated} when neither is an ancestor of
the other. The \emph{distance} between two structures is the length
of the shortest path between them in the graph obtained from $G$ by
forgetting the directions of its edges.

\paragraph{Corpora.} For each concept, Llama-3.1-8B-Instruct writes up
to 90 texts of at most 200 words that use the concept, varying the
subfield, the difficulty and the register, and is asked not to name
it; 38\% of the texts still name their concept. The texts about the 18
instances never name the algebraic, graph or topological structures
related to the instance. A judge drops the texts that do not actually
use the concept, and each corpus keeps 35--90 texts. The same
generator writes one corpus for each of 48 emotions, which never names
the emotion (\cref{app:emotions}). We also build smaller corpora from
the Wikipedia articles of 94 concepts (\cref{app:corpora}).

\paragraph{Concept vectors.} A text is represented by the activation
at its last token after a given layer \citep{azaria2023internal,marks2024geometry}.
Following the contrastive mean-difference recipe
\citep{sofroniew2026emotion,chen2025persona,lindsey2025introspection},
the \emph{concept vector} of a concept $c$ at layer $\ell$ is
$v_c = \mu_c - \bar\mu$, where $\mu_c$ is the mean activation on
$c$'s corpus and $\bar\mu$ the average of the $\mu_{c'}$ over all
concepts $c'$. This \emph{centering} removes a large component that
vectors built against a single baseline corpus share
(\cref{app:variants}). We call \emph{direction} any unit vector, that
is, any vector of euclidian norm 1, and the direction of a nonzero
vector the unit vector proportional to it. The \emph{score} $s_c(t)$
of a text $t$ on $c$ is the projection of its activation on the
direction of $v_c$, rebuilt without $t$ when $t$ is in $c$'s corpus.
Results are reported over the plateau of layers on which concepts are
best separated, from about a quarter of the depth to the
second-to-last layer (\cref{sec:storage}).

\paragraph{Word statistics.} The \emph{word-statistics
representation} of a text is the vector of TF-IDF weights of its words
and pairs of consecutive words, without stop words, projected on 512
directions by latent semantic analysis \citep{deerwester1990lsa}, so
that texts on the same subject get nearby representations even
without a common word (\cref{app:wordstats}). Building concept vectors
and scores from it in the same way as from activations gives
\emph{word-statistics concept vectors} and \emph{scores}, and every
experiment that uses concept vectors is also done with them. We also
regress the measures of the model on those of word statistics, and
recompute the results on the residuals.

\paragraph{Writing vectors.} The \emph{writing vector} of a token $v$
at layer $\ell$ is the direction that, added after layer $\ell$, most
raises the logit of $v$, to first order. We obtain it from the
Jacobian lens \citep{anthropic2026jlens}, the mean $J_\ell$, over
generic texts, of the Jacobian of the map from the state of the
residual stream after layer $\ell$ to the final state: it is the
direction of $J_\ell^{\top}(g\odot W_U[v])$, where $W_U[v]$ is the
output embedding of $v$ and $g$ the weights of the final normalization
(\cref{app:writing}). The writing vector of a concept is that of its
name when the name is a single token, and that of the first token of
its name otherwise, which is only a proxy.

\paragraph{Steering.} We steer by activation addition
\citep{turner2023steering,rimsky2024steering}: while the model
continues a prompt, for at most 400 tokens, we add
$\alpha\,\bar n_\ell\,u$ to the residual stream after layer $\ell$ at
every position, where $u$ is a direction, $\alpha$ the
\emph{intensity}, and $\bar n_\ell$ the median norm of the activations
at this layer. Each \emph{steered completion} is compared with the
\emph{unsteered completion} generated from the same prompt and seed.
The \emph{readout} of a concept is the projection on its concept
vector of the mean activation of the completion, passed again through
the model without the prompt and without steering. We report its
change, averaged over prompts, in standard deviations of the unsteered
readouts. A \emph{judge} (Qwen2.5-72B-Instruct) sees a steered
completion and its unsteered completion in random order, and says
which is more about the concept (\cref{app:steer-methods}).

\paragraph{Models.} We use DeepSeek-R1-Distill-Qwen-7B (28 layers) and
DeepSeek-R1-Distill-Qwen-32B (64 layers) \citep{deepseekai2025r1}
throughout, and call them R1-Qwen-7B and R1-Qwen-32B. In
\cref{sec:geometry}, we also use DeepSeek-R1-Distill-Llama-8B
(R1-Llama-8B), which starts from a base model of another family, and
Qwen2.5-Math-7B, the base model of R1-Qwen-7B before reasoning
training. A result reported at a single layer is at layer 16 of
R1-Qwen-7B or layer 36 of R1-Qwen-32B, both in the plateau. Intervals
are 95\% bootstrap intervals, which resample structures for statistics
over pairs of structures (\cref{app:ci}).

\section{Steering validates mathematical concept vectors only weakly}
\label{sec:writing}

Concept vectors have been validated by steering
\citep{chen2025persona,sofroniew2026emotion}: adding a concept vector
to the activations makes the model write more about the concept, which is
taken as a sign that the vector is related to its concept. For mathematical concepts, this happens only
weakly. Steering has an effect only in a small window of
intensities. Within this window, it changes the frequency of the name
of a concept in texts where the concept is already relevant, and
brings it into no other text (\cref{sec:steering}). This is consistent with a second result:
the concept vector is nearly
orthogonal to the writing vector of the concept, unlike for emotions
(\cref{sec:readwrite}).

\subsection{Mathematical concept vectors steer only in a small window of intensities}
\label{sec:steering}

We expect steering along the concept vector of a concept to have
little effect on a text already about this concept, because
mathematical concepts are rigid, and little effect on an unrelated
text for the same reason, while emotions can always be expressed with
more intensity. We expect the most effect when steering along the
concept vector of a parent of the concept the text is about. We do so
for twelve pairs of concepts, with intensities from $-2$ to $-0.005$
and from $0.005$ to $2$, and add the concept vector of each of 53
concepts on 40 non-conceptual mathematical prompts, such as ``How many
seconds are in a week?'', at layer 16 of R1-Qwen-7B and layer 36 of
R1-Qwen-32B, with random directions as controls. Steering too strongly
makes the model write incoherent text. We measure this by the
\emph{score increase}: the negative log-likelihood per token of a
steered completion under the unsteered model, minus that of its
unsteered completion. We calibrate the judge on unsteered completions
into which we insert 1, 2 or 4 sentences from the corpus of the parent
or of an unrelated concept (\cref{app:steering}).

\begin{figure*}[t]
\centering
\includegraphics[width=\textwidth]{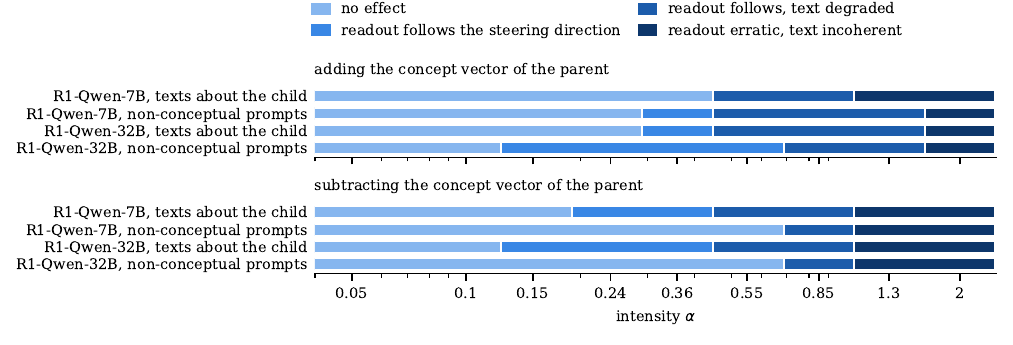}
\caption{Regimes of steering along the intensity $\alpha$
(\cref{sec:steering}), for each model and type of text, when adding
(top) and subtracting (bottom) the concept vector of the parent.
The readout of the parent follows the steering direction when it
moves by at least 0.1 standard deviations in this direction, and the
text is degraded when the score increase is 0.2 or more.
\Cref{app:regimes} defines the regimes.}
\label{fig:steering_regimes}
\end{figure*}

\paragraph{Results.} Only in a narrow window of intensities
(\cref{fig:steering_regimes}) does adding the concept vector of the
parent raise its readout on texts about the child, and subtracting it
lower this readout. The window goes from $\alpha = 0.15$ to 0.55 on
R1-Qwen-32B, and from 0.24 (subtracting) or 0.36 (adding) to 0.55 on
R1-Qwen-7B, and over it the score increase reaches 0.24--0.26 on
R1-Qwen-32B and 0.45--0.55 on R1-Qwen-7B. Above it, the readout of the
parent no longer follows the steering direction and the completions
become incoherent (\cref{app:regimes}).

Within the window, the text changes much less than the readout. On
non-conceptual prompts, at $\alpha = 0.3$, the concept vector moves
its own concept in the steering direction for 48 of the 53 concepts
on R1-Qwen-32B (36 on R1-Qwen-7B), but the name of the concept stays
at most 0.01 occurrences per 100 words on R1-Qwen-7B, with or without
steering. One inserted sentence that never names the parent moves its
readout by $+0.15$ standard deviations, and the judge detects it
($+0.33$). Steering moves the readout further ($+0.21$ at $\alpha =
0.55$), and the judge does not see it ($+0.00$ to $+0.01$). On texts
about the child concept, on R1-Qwen-7B, adding the parent raises the
frequency of its name from 0.51 to 0.69 and 0.73 occurrences per 100
words at $\alpha = 0.36$ and 0.55, and subtracting it lowers it to
0.43 and 0.33. The judge sees the removal ($-0.18 \pm 0.04$ at $\alpha
= -0.36$, against $-0.11$ for random directions), and not the
addition.

\paragraph{Interpretation.} Steering along a mathematical concept
vector moves the readout of the concept within the window, but its
effect on the content of the text is small, and we found it only where
the concept is already relevant. Steering can thus be considered at
best a weak validation of mathematical concept vectors. For
comparison, \citet{sofroniew2026emotion}, who measure intensities in a
similar unit, report large effects of emotion vectors on behaviour
from an intensity of 0.05, such as a blackmail rate going from 22\% to
72\% in one scenario, while at this intensity mathematical concept
vectors leave even the readout unchanged in our models. The models,
the layers and the measured effects differ, so this comparison is only
indicative. No variation of the intervention we tried makes the judge
find the text more about the concept (\cref{app:sweep}).

\subsection{Mathematical reading and writing vectors are nearly orthogonal}
\label{sec:readwrite}

Steering along the concept vector of an emotion makes the model
produce text in line with the emotion \citep{sofroniew2026emotion},
which suggests that this concept vector is related to the writing
vector of the name of the emotion. We compare three groups of
concepts: 48 emotions, mathematical concepts named by a common word
(ring, field, average, \dots), whose names have other meanings, and
the other mathematical concepts. The main comparison is between
concepts whose name is a single token: 43 of the 48 emotions and 15 of
the 136 mathematical concepts. The emotion corpora never name their
emotion, while 38\% of the mathematical texts name their concept,
which should, if anything, bring mathematical concept vectors closer
to their writing vectors.

\paragraph{Cosines.} At the steering layer, the cosine between a
mathematical concept vector and its writing vector lies between
$-0.06$ and $+0.11$ on R1-Qwen-7B and between $-0.08$ and $+0.09$ on
R1-Qwen-32B. Most of these cosines are of the order of those of two
unrelated directions (about $\pm 0.017$ and $\pm 0.014$), though
mostly positive (median $+0.02$ and $+0.01$). For emotions, the cosine
is $+0.06$ and $+0.09$ on average, several times larger.

\paragraph{Identification rate.} The relevant reference for the
cosine of a concept vector with its writing vector is its cosines with
the writing vectors of the other concepts of its group. We draw 200
random sets of 12 concepts of the same group, and call
\emph{identification rate} the proportion of concepts whose writing
vector has, among the 12, the highest cosine with their concept vector
($1/12 \approx 0.08$ by chance). There are two natural expectations.
The writing vector of a name with several meanings writes the word in
all its senses, so concepts with a common name would be identified
less well than technical ones. Emotions vary along a few shared axes,
such as positive or negative, so their names could be identified along
these axes alone. The identification rate of emotions is well above
chance from the middle layers on (\cref{tab:readwrite}), and removing
the top two principal components does not lower it (R1-Qwen-32B, layer
50: 0.86 before and after). Mathematical concepts with a common name
come between emotions and technical concepts, and all three groups are
identified well only in the last layers. Restricted to single-token
names, the identification rate is 0.78 for emotions against 0.28 for
mathematical concepts (R1-Qwen-32B, layer 36), and 0.75 against 0.36
(R1-Qwen-7B, layer 16).

\paragraph{Steering.} We compare how much the concept vector, the
writing vector and a random direction raise the probability of the
name. We add each of them at the steering layer while fixed texts pass
through the model, rank all tokens by how much it raises their
log-probability, and report the rank of the first token of the name
(\cref{app:readwrite}). The writing vector puts the name at median
rank 1 in every group, random directions at median ranks between
3{,}500 and 106{,}000, and the concept vector at 46--60 (R1-Qwen-7B)
and 20--40 (R1-Qwen-32B) for emotions, and much further for
mathematical concepts (\cref{tab:readwrite}). Across 48 mathematical
concepts, the rank follows the cosine with the writing vector: the
Spearman correlation between the cosine and the logarithm of the rank
is $-0.75$ on R1-Qwen-7B and $-0.85$ on R1-Qwen-32B at $\alpha =
0.55$, and the ten concepts with the largest cosines, comparable to
those of emotions, put their name at median rank 13 and 131.

\begin{table}[t]
\centering
\small
\caption{Reading and writing vectors (\cref{sec:readwrite}).
Identification rate, among 12 concepts, from the cosine between a
concept vector and the writing vectors of the names (R1-Qwen-32B, chance 0.08), at
three layers, and median rank of the name when steering with the
concept vector. The first three rows include names of more than
one token, with the first-token proxy. The last two rows (1 token)
keep only single-token names. Ranks are for R1-Qwen-7B and
R1-Qwen-32B, at $\alpha = 0.36$ and 0.55.}
\label{tab:readwrite}
\setlength{\tabcolsep}{3pt}
\definecolor{tabhead}{gray}{0.92}
\begin{tabular}{>{\columncolor{tabhead}}l|ccc|cc}
\rowcolor{tabhead}
 & \multicolumn{3}{c|}{identification rate} & \multicolumn{2}{c}{median rank} \\
\rowcolor{tabhead}
 & L20 & L40 & L60 & 7B & 32B \\
\hline
emotions & 0.56 & 0.76 & 0.87 & 46--60 & 20--40 \\
common-word math & 0.14 & 0.32 & 0.83 & 160--200 & 1.9k--3.7k \\
other math & 0.06 & 0.19 & 0.64 & 1.5k--3.8k & 16k--29k \\
\hline
emotions, 1 token & 0.58 & 0.81 & 0.88 & --- & --- \\
math, 1 token & 0.14 & 0.27 & 0.82 & --- & --- \\
\end{tabular}
\end{table}

\paragraph{Interpretation.} For emotions, the concept vector is
clearly aligned with its writing vector, though far from equal to it.
For mathematical concepts, it is nearly orthogonal to it, and the
small alignment that remains is specific but too small for the concept
vector to write the name. Neither of the expectations we formulated
holds, so neither explains the gap between emotions and mathematical
concepts. A linear approximation explains why the rank of the name
follows the cosine. Adding the concept vector at intensity $\alpha$
moves the activations along the writing vector only by $\alpha$ times
their cosine. With cosines of 0.01--0.02, the concept vector needs an
intensity 50 to 100 times larger than the writing vector to push the
name as much, far beyond the intensities at which the text becomes
incoherent. For emotions, average cosines of $+0.06$ and $+0.09$ reduce
this factor to 11--18. Conversely, the writing vector of a concept
reads it only weakly, about a paragraph before the model writes its
name (\cref{sec:writeread}).

\section{Concept vectors reflect the symmetric part of the inclusion graph}
\label{sec:geometry}

For emotions, the relations between concepts are reflected in the
geometry of their concept vectors \citep{sofroniew2026emotion}. In
mathematics, they are given by definitions. We ask whether concept
vectors separate the texts about their concept from the other texts,
respect the inclusions between concepts, and compose as definitions
do, on the four models of \cref{sec:method}.

\subsection{Concept vectors separate concepts, less well than word statistics}
\label{sec:storage}

We call \emph{separation AUC} of a concept $c$ at a layer the AUC of
the score $s_c$ between the texts of $c$'s corpus and all the other
texts, and average it over concepts. It is 0.49--0.50 on the input
embeddings, rises over the first quarter of the depth, and stays on a
plateau up to the second-to-last layer (0.72--0.79), over which the
similarities between concept vectors barely change (correlation
0.98--0.99 between layers; \cref{app:layers}). Concept vectors thus
separate concepts, and the network builds this separation. Word statistics separate them better,
with a separation AUC of 0.98. Among texts that word statistics score
about equally, the layers of the plateau still separate concepts at
0.59--0.64, below word statistics themselves in the same comparison
(0.86; \cref{app:separation}). Separation thus cannot show that concept
vectors represent their concept beyond the vocabulary of the texts
about it.

\subsection{Concept vectors reflect the distance in the inclusion graph and miss inclusion}
\label{sec:inclusion}

Every instance of a structure is an instance of each of its
ancestors, so a text about a structure is also about its ancestors.
We define the \emph{ratio} $r(c\to c')$ of two concepts as the mean
projection of the texts of $c$ on $v_{c'}$, divided by the same mean
for the texts of $c'$ (\cref{app:inclusion}). It is 1 when the texts
of $c$ score on $c'$ as high as those of $c'$, and 0 when they score
as the average text. An \emph{ancestor pair} is a pair
$(c, c')$ where $c'$ is an ancestor of $c$. \emph{Containment}: the
texts of a concept should score higher on its ancestors than on
unrelated structures. Since ancestor pairs are on average closer in
$G$, the \emph{containment AUC} is the share of comparisons between an
ancestor pair and an unrelated pair at the same distance, of at least
3, in which the ancestor pair has the higher ratio.
\emph{Directionality}: a text about $c'$ is in general not about its
descendant $c$, so we should have $r(c\to c') > r(c'\to c)$, and the
\emph{directionality} is the share of ancestor pairs for which this
holds. We use 97 structures and 520 ancestor pairs, at layers 12, 16
and 20 of the four models, and compute the same measures on word
statistics and on the residuals of the ratios of the model after
regression on those of word statistics.

The ratio falls steadily with the distance in $G$, from a concept to
its ancestors and between unrelated structures in both orders
(R1-Qwen-7B, layer 16: 0.42 for ancestor pairs at distance 3, 0.26
at distance 5).
The containment AUC is 0.49--0.55 and the directionality 0.54--0.59 on
every model and layer, and every interval includes chance
(\cref{tab:containment}). Word statistics give the same decay and the
same results at chance (containment AUC 0.50, directionality 0.47).
The residuals no longer decrease with the distance, and their
containment AUC and directionality stay at chance (0.50--0.56 and
0.53--0.58). Dropping the texts that name an ancestor, using the
Wikipedia corpora, or using the estimator of \citet{park2025geometry}
changes none of these results (\cref{app:inclusion}).

\begin{table}[t]
\centering
\small
\caption{Containment AUC and directionality (\cref{sec:inclusion}),
97 structures, 520 ancestor pairs, layer 16, concept-level bootstrap.
Resid.: containment AUC on the residuals of the ratios of the model
after regression on the word-statistics ratios, which is not defined
for word statistics themselves.}
\label{tab:containment}
\setlength{\tabcolsep}{2.5pt}
\begin{tabular}{lccc}
\toprule
 & Containment & Resid. & Direction \\
\midrule
R1-Qwen-7B & 0.55 [0.43, 0.66] & 0.56 & 0.56 [0.47, 0.65] \\
R1-Qwen-32B & 0.50 [0.38, 0.63] & 0.51 & 0.57 [0.48, 0.66] \\
R1-Llama-8B & 0.51 [0.39, 0.64] & 0.53 & 0.57 [0.49, 0.65] \\
Qwen2.5-Math-7B & 0.52 [0.41, 0.64] & 0.54 & 0.54 [0.45, 0.62] \\
Word statistics & 0.50 [0.39, 0.62] & --- & 0.47 [0.38, 0.58] \\
\bottomrule
\end{tabular}
\end{table}

Concept vectors thus reflect the symmetric part of the inclusion
graph, the distance between concepts, and this part is captured by the
word statistics of the texts. They do not reflect its asymmetric
part: the ratio is no higher for ancestors than for unrelated
structures at the same distance, and no higher from a concept to its
ancestor than in the other direction.

\subsection{Concept vectors do not compose like definitions}
\label{sec:composition}

The definition of a structure is built from other concepts, its
\emph{ingredients}: a monoid is a semigroup with an identity element.
We look for two forms of linear composition, weaker than additivity,
since macro-interpretability uses concept vectors through linear
operations (\cref{app:composition}). First, for 14 structures, the
closest linear combination of the vectors of its ingredients explains
$R^2 \approx 0.2$ of the vector of a structure, against 0.12--0.17 with
random concepts. Second, we call \emph{offset} the difference between
the vectors of two structures whose definitions differ by one
requirement, such as having inverses. Over 21 such pairs, the offsets
of pairs that add the same requirement have a mean cosine of
0.03--0.06, against 0.005--0.014 for different requirements, and word
statistics align them more (0.17). Concept vectors thus satisfy
neither form of linear composition.

\medskip
\noindent In summary, word statistics do as well as concept vectors or
better wherever we compare them, including on the model's own chains
of thought (\cref{sec:cot}).

\section{The model represents the asymmetric part that concept vectors miss}
\label{sec:aboutness}
\label{sec:where}

\Cref{sec:geometry} found that concept vectors reflect how close two
concepts are in the inclusion graph, but not which one includes the
other. To tell whether this comes from the model or from the notion of
concept vector, we compare the model and concept vectors on whether an object $x$ is an
instance of a structure $c$. The informative cases are the \emph{near
misses}, where $x$ is an instance of a parent of $c$ without being an
instance of $c$, such as the symmetric group $S_3$ for abelian group.
We use the structures of three \emph{families} (algebra, graphs and
topology) and the 18 instances of the bank (\cref{app:asymmetry}),
with three kinds of texts: the \emph{instance texts}, which are the
corpora of the instances and never name a structure of their family;
2{,}604 templated \emph{statements} ``$x$ is (not) a $c$'', read at
their last token, of which 264 are near misses; and 118
\emph{questions} ``Is $x$ a $c$? Answer yes or no.'', read at the last
token of the name of $x$, the \emph{object token}, which comes before
$c$ is named and is where attributes of a subject are retrieved
\citep{meng2022rome,geva2023dissecting}. We use six \emph{minimal
contrasts}, each a structure with two instances of one of its parents,
only one of which is an instance of the structure ($\mathbb{Z}$ and
$S_3$ for abelian group).

\subsection{The model represents which objects are instances of a structure}
\label{sec:probe}

We train a \emph{probe}, a linear classifier, to predict from the
activation at the last token of a statement whether it is true. For
each structure, the probe is trained on the statements about the
other structures, so that it can only succeed if the model represents
membership in the same way for all structures. Its AUC between near
misses and true statements, which requires telling a structure from
its parent, is about 0.85 on R1-Qwen-7B and 0.98 on R1-Qwen-32B, and
the same with one instance held out. It gets 66 and 68 of the 72
comparisons of the minimal contrasts right on R1-Qwen-7B (layers 20
and 24), and all 72 on R1-Qwen-32B. It reads the pair of the object
and the structure, and not the object alone (\cref{app:probe}). The
same probe on word counts reaches 0.58 with one instance held out and
0.20 with one structure held out. The model thus represents
membership, including its asymmetry, linearly and beyond the word
statistics of the statements, through a truth direction shared by
all structures \citep{burns2022discovering,azaria2023internal,marks2024geometry,li2023iti}.

\subsection{Concept vectors miss membership, and variants of them capture it}
\label{sec:directions}

We score the objects of a family on the concept vector of a structure
$c$ in three ways: each instance by the mean score of its texts, each
positive statement ``$x$ is a $c$'', and each of 77 objects, 59 of
them outside the bank, by the score of its object token in the
questions (\cref{app:scoring}). The AUC between the instances of $c$
and the other objects of the family, averaged over structures, is
0.44--0.67 (\cref{tab:constructions}), no better than word statistics
(0.61--0.64), and on the instance texts the concept vector gets at
most 2 of the 6 minimal contrasts right. The texts about $S_3$ even
score significantly above those about $\mathbb{Z}$ on abelian group
(\cref{sec:topic}).

We then vary the definition of the concept vector of $c$. The
\emph{statement direction} is the direction of the difference between
the mean activation of the positive statements ``$x$ is a $c$'' with
$x$ an instance of $c$ and the same mean with $x$ not an instance. The
\emph{question direction} is the direction of the same difference at
the object token of the questions. Both are built without the statements and
questions about the object being scored. The question direction reaches
0.82--0.89 on the questions and 0.73--0.89 on the instance texts, and
the statement direction 0.80--0.91 on the statements of R1-Qwen-32B,
though neither gets more than 2 of the 6 minimal contrasts right on
the instance texts. The failure of the concept vector to capture
membership thus comes from the choices made within its form, a
difference between mean activations: which texts are averaged, at
which position they are read, and what is subtracted.

\begin{table}[t]
\centering
\small
\caption{AUC between instances and non-instances of a structure, averaged over structures, for the concept vector, the statement direction and the question direction
(\cref{sec:directions}), objects held out, evaluated on instance texts that
never name a structure, on positive statements, and on the object token
of the questions.}
\label{tab:constructions}
\begin{tabular}{lccc}
\toprule
Evaluated on & Concept & Statement & Question \\
\midrule
\multicolumn{4}{l}{R1-Qwen-7B} \\
\quad instance texts & 0.48--0.61 & 0.55--0.67 & \textbf{0.85--0.89} \\
\quad questions & 0.45--0.49 & 0.58--0.64 & \textbf{0.84--0.87} \\
\multicolumn{4}{l}{R1-Qwen-32B} \\
\quad instance texts & 0.54--0.59 & 0.55--0.76 & 0.73--0.89 \\
\quad statements & 0.45--0.67 & \textbf{0.80--0.91} & 0.66--0.73 \\
\quad questions & 0.44--0.52 & 0.60--0.70 & \textbf{0.82--0.89} \\
\bottomrule
\end{tabular}
\end{table}

\section{Discussion}
\label{sec:discussion}

\Cref{sec:directions} shows that a vector defined like the concept
vector, as a difference between mean activations, captures more of
the structure of mathematics with other choices of texts, position and
reference. If these choices follow a principle that
applies to any concept, they would improve macro-interpretability in
other domains as well.

\paragraph{Parameter choices.} Our choices, texts about the concept
read at their last token with the mean over all corpora subtracted,
suit concepts that are spread over a text, as an emotion usually is.
What a mathematical text says about a structure is more localized, at
the tokens of the objects that instantiate it and of the claims made
about them, where the question direction reads. This suggests choosing the texts and the position according to what
instantiates the concept: the text itself for an emotion, the token of
the object for a property of objects, and possibly the tokens of both
objects for a relation between them. Since the concept vector differs
from the question direction in its texts, its position and its
reference, a direct test would build, for concepts of several kinds,
the vectors given by each combination of these choices.

\paragraph{Why concept vectors of emotions write and those of
mathematical concepts do not.} A concept vector can write only what
distinguishes the texts of its concept in the words most likely to
extend them. Texts in different emotions differ largely by their tone,
which these words reflect, while in mathematical texts these words are
more often connecting words, unaffected by the concept. Following this hypothesis, techniques with a
signature vocabulary (induction and its ``base case'') should have
concept vectors that write the concepts, and steering should write
more where the next word could name an object or the concept.

\section{Limitations}
\label{sec:limitations}

Our experiments could be extended to
other families and models larger than 32B, natural texts in place
of generated corpora and templated statements, writing vectors for
names of several tokens, other judges, and the steering of emotion
vectors under our protocol. The regimes of
\cref{fig:steering_regimes} rest on thresholds that we chose, and
several negative results rest on small samples (14 compositions, 21
pairs adding a requirement, 6 minimal contrasts, 12 parent--child
pairs).


\section{Conclusion}
\label{sec:conclusion}

In mathematics, concepts are related by a more definite structure
than emotions, which allows us to draw precise limits of
macro-interpretability methods. Mean-difference concept vectors
pass the usual validation by steering only weakly, as their small
cosine with the writing vector of the name predicts. The structure of mathematics provides a finer
validation method. Concept vectors preserve the symmetric part of
the inclusion graph, corresponding to the distance in this graph,
which word statistics already capture. They do not reflect its asymmetric part,
corresponding to the inclusion of a concept in another and to the
membership of an instance in a type of structure. The model itself reflects
this asymmetry, so the fact that concept vectors do not reflect it
comes from their definition.
Concept vectors thus give an incomplete account of the concepts they
stand for, and an exact structure, as in mathematics, reveals which
part they miss. Concepts without such a structure, such as emotions,
may have aspects that concept vectors miss as well, without any way
of seeing it. The preservation of a known structure can therefore
serve as a common yardstick for notions of concept vector, and guide
the refinement of the current one.

\bibliography{biblio}
\bibliographystyle{icml2026}

\appendix
\section{Emotions}
\label{app:emotions}

The comparison of \cref{sec:readwrite} uses 48 emotions: joy, anger,
fear, sadness, disgust, surprise, pride, shame, guilt, envy, hope,
relief, love, hatred, loneliness, nostalgia, gratitude, boredom,
anxiety, calm, contentment, excitement, frustration, embarrassment,
awe, admiration, contempt, curiosity, confusion, despair,
determination, disappointment, regret, resentment, sympathy,
tenderness, trust, worry, amusement, longing, panic, serenity, shyness,
triumph, horror, irritation, melancholy and jealousy. The corpus of
each emotion is written by Llama-3.1-8B-Instruct, with three texts for
each combination of a register (short story, diary entry, letter to a
friend, dialogue, first-person monologue) and a setting (at work, in a
family, while travelling, at school, during a sports match, in a
hospital). The prompt asks to convey the emotion through the
situation, the thoughts and the actions of the characters, and texts
that name the emotion are dropped. Concept vectors are computed as for
mathematical concepts, with the mean over the 48 emotion corpora
subtracted.

\section{Material and methods}
\label{app:methods}

This appendix gives the details of the material and the methods that
\cref{sec:method} introduces: the units of each experiment, the
concept bank and its inclusion graph, the corpora, the scores, the
word-statistics representation, the confidence intervals, steering
and its measures, the Jacobian lens and writing vectors, and the
construction of the concept vectors.

\subsection{Units of each experiment}
\label{app:units}

The bank has 201 concepts, but no experiment uses all of them. Each experiment
uses the subset it needs, and its sample size is the number
of these units, which is often much smaller than the bank.
\Cref{tab:units} lists the units of each experiment, the main text section
that reports it, and the unit we resample for intervals.

\begin{table}[h]
\centering
\small
\caption{Units of each experiment.}
\label{tab:units}
\begin{tabular}{p{0.34\linewidth}p{0.52\linewidth}}
\toprule
Experiment & Units \\
\midrule
Separation AUC, conditional separation AUC (\cref{sec:storage}) & 136 concepts (every
kind except the structures used only by the inclusion experiment) \\
Containment, directionality (\cref{sec:inclusion}) & 97 structures, 520
ancestor pairs; resampled by structure \\
Wikipedia corpora (\cref{sec:inclusion}) & 94 concepts \\
Composition (\cref{sec:composition}) & 14 definitions with hand-listed
ingredients \\
Added requirements (\cref{sec:composition}) & 21 pairs adding one of 6 requirements \\
Relations (\cref{sec:composition}) & 6 structures $\times$ 4
relations, 22 filled cells \\
Instance texts (\cref{sec:topic}) & 18 instances; 6 minimal contrasts \\
Statements (\cref{sec:topic,sec:probe}) & 2{,}604 statements over 18
instances \\
Chains of thought (\cref{sec:cot}) & 360 chains; 80 competition and 200
MATH problems \\
Steering (\cref{sec:steering}) & 12 parent--child pairs; 53 concepts on
non-conceptual prompts \\
Reading vs.\ writing (\cref{sec:readwrite}) & 48 emotions; mathematical
concepts in two groups; 12 steering pairs \\
\bottomrule
\end{tabular}
\end{table}

\subsection{Concept bank and inclusion graph}
\label{app:bank}

\paragraph{Concept bank.} The 201 concepts of the bank are comparable
in number with the 171 emotions studied by \citet{sofroniew2026emotion}.
Each of them has a
name, a kind, a field of mathematics and a one-sentence description,
and the bank records relations between concepts: each instance lists
the structures it belongs to, and each structure its direct parents.
The 101 structures include a few basic notions, such as set and
function, and are listed below by field.

\paragraph{Inclusion graph.} The \emph{inclusion graph} has
mathematical definitions as vertices, and an edge from a definition $D$
to a definition $D'$ when $D'$ generalizes $D$, in the sense that every
object that satisfies $D$ also satisfies $D'$. For instance, there is
an edge from the definition $D$ of a field to the definition $D'$ of a
ring, since every field is a ring.
We denote by $G$ the graph on the structures of the concept bank whose
edges are the direct generalizations, those that do not follow from
others by composition. In the concept bank, each structure lists its direct
parents, which give the edges of $G$. The \emph{ancestors} of a
structure are the structures reached from it by a directed path in
$G$, that is, the structures of the bank that generalize it. A
structure is a \emph{descendant} of each of its ancestors, and two
structures are \emph{unrelated} when neither is an ancestor of the
other. $G$ has six connected components.
The largest joins algebra, topology and analysis, through structures
such as topological groups and normed spaces, and the others cover
graphs, polygons, order theory, category theory and measure theory.
$G$ is not a tree, since 22 structures have several parents (a field is
both an integral domain and a division ring). The \emph{distance}
between two structures is the length of the shortest path between them
in the graph obtained from $G$ by forgetting the directions of its
edges.

The 101 structures, by field:

\paragraph{Algebra (45).} set, function, polynomial, semigroup, monoid, group, abelian group, ring, integral domain, division ring, field, module, vector space, unique factorization domain, principal ideal domain, Euclidean domain, Lie group, commutative monoid, quasigroup, loop, cyclic group, finite group, simple group, solvable group, nilpotent group, finite $p$-group, free group, topological group, semiring, commutative ring, local ring, Noetherian ring, Dedekind domain, algebraically closed field, finite field, ordered field, associative algebra over a field, Lie algebra, projective module, free module, category, groupoid, additive category, abelian category, topos.

\paragraph{Combinatorics (11).} graph, tree, forest, bipartite graph, planar graph, connected graph, regular graph, cubic graph, complete graph, Eulerian graph, Hamiltonian graph.

\paragraph{Geometry and topology (23).} topological space, metric space, quadrilateral, parallelogram, rectangle, rhombus, square, manifold, Hausdorff space, uniform space, compact space, compact Hausdorff space, connected space, path-connected space, smooth manifold, Riemannian manifold, complex manifold, polygon, triangle, isosceles triangle, equilateral triangle, right triangle, regular polygon.

\paragraph{Analysis and probability (13).} random variable, normed vector space, inner product space, Banach space, Hilbert space, compact metric space, Banach algebra, C$^*$-algebra, complete metric space, measurable space, measure space, finite measure space, probability space.

\paragraph{Order theory (9).} preorder, partial order, total order, well-order, lattice, distributive lattice, Boolean algebra, complete lattice, directed set.

The 14 properties and operations used in definitions are operation,
edge relation, scalar action, open sets, distance, associativity,
identity element, inverses, commutativity, distributivity, zero
divisors, connectedness, acyclicity and triangle inequality.

\subsection{Corpora}
\label{app:corpora}

For each concept, Llama-3.1-8B-Instruct, a model of a
different family from most of the models we study, writes up to 90 texts of at most 200 words that use the concept,
and is asked not to name it. The prompt varies three factors, with two
texts for each combination: the subfield (algebra,
combinatorics, geometry, number theory or probability), the difficulty,
stated only as easy, medium or hard and left to the generator's
interpretation, and the register, that is, the kind of text (textbook
exposition, competition-style solution or step-by-step tutorial). The generator does not always comply, and 38\% of the texts
name their concept. For the 18 instances, the prompt forbids the names
of the algebraic, graph or topological structures, according to the
instance (for $\mathbb{Z}$: group, abelian, ring, field, and so on), and a text that still contains one is
regenerated, so that these texts never name such a structure. A judge drops the texts that do not actually use
the concept. Each corpus keeps 35--90 texts. For the comparison of \cref{sec:readwrite}, the same
generator writes one corpus for each of 48 emotions (\cref{app:emotions}),
with three texts for each combination of five registers and six
settings. The emotion is conveyed through situations, thoughts and
actions, and texts that name it are dropped, so that these corpora
never name their emotion. A mention audit records
which other concept names each text contains. We generate the main
corpora because generation controls what the texts contain: the
texts vary along the same factors for every concept. Natural sources cover fewer concepts, since
instances and narrow constructions rarely have an encyclopedia article
of their own, and their paragraphs mix the concept with its history,
applications and notation. To make sure that our results do not depend on
text generation, we also build smaller corpora from the paragraphs of
each concept's English Wikipedia article, for the 94 concepts whose
article has at least 10 paragraphs.

\subsection{Scores}
\label{app:scores}

\Cref{sec:method} defines the score of a text on a concept as the
projection of its activation on the direction of the concept vector,
rebuilt without the text when the text belongs to the corpus of the
concept. We give the exact definition here.

For a layer $\ell$ and a text $t$, let $a^{(\ell)}(t)$ be the
activation of the last token of $t$ after layer $\ell$, and for a
concept $c$, let $C_c$ be its corpus. For $t \in C_c$, the concept
vector of $c$ rebuilt without $t$ is
\[
v_c^{(\ell),-t} := \Big( \frac{1}{|C_c|-1}
\sum_{t' \in C_c \setminus \{t\}} a^{(\ell)}(t') \Big) - \bar\mu^{(\ell)},
\]
where the reference $\bar\mu^{(\ell)}$ over all the corpora is kept,
because the contribution of $t$ to it is negligible. Indeed, the weight
of $t$ is $1/|C_c|$ in the mean over $C_c$, and only
$1/(n|C_c|)$ in the reference, for $n$ concepts. The first weight needs to be removed because $a^{(\ell)}(t)$ is one of
the terms of the mean over $C_c$, so that the product $\langle a^{(\ell)}(t), v_c^{(\ell)}
\rangle$ contains the term $\|a^{(\ell)}(t)\|^2/|C_c|$. This term is large
compared with the terms $\langle a^{(\ell)}(t), a^{(\ell)}(t')
\rangle/|C_c|$ of the other texts $t'$ of $C_c$, because in high
dimension the parts of two activations that the texts do not share are
nearly orthogonal, so that it inflates the score even for random
activations. Writing $\hat v := v/\|v\|$ for a vector $v$, the
\emph{score} of $t$ on $c$ at layer $\ell$ is
\[
s_c^{(\ell)}(t) :=
\begin{cases}
\langle a^{(\ell)}(t), \hat v_c^{(\ell)} \rangle & \text{if } t \notin C_c,\\
\langle a^{(\ell)}(t), \hat v_c^{(\ell),-t} \rangle & \text{if } t \in C_c,
\end{cases}
\]
so that a text never contributes to the vector it is scored on. As for
the concept vector, we drop $\ell$ from these notations when the layer is clear from
the context. The score of a text from another corpus measures how
much of $c$ this text carries. The scores of $c$'s own texts give the
reference level of $c$, against which the other scores are compared.

\subsection{Word statistics}
\label{app:wordstats}

The word-statistics
representation $a^{\mathrm{ws}}(t)$ of a text $t$ is built from
classical statistics of the occurrences of its words, as follows.
First, the \emph{stop words} of $t$ are removed. These are words such
as \emph{the}, \emph{of} or \emph{is}, which carry little content,
taken from a standard list. The \emph{terms} of $t$ are then its
remaining words and the pairs of words that are consecutive once the
stop words are removed (for instance, \emph{the ring of integers} yields the pair
\emph{ring integers}). Only the terms, words or pairs, that appear
in at least two texts of the corpora are kept.
Let $V$ be the set of kept terms and $N$ the number of texts. For a
term $u \in V$, let $n_u(t)$ be its number of occurrences in $t$ and
$d_u$ the number of texts that contain it. The \emph{TF-IDF weight} of
$u$ in $t$ is
\[
x_u(t) := \Big(1 + \log\big(n_u(t)\big)\Big)\bigg(1 + \log\Big(\frac{1+N}{1+d_u}\Big)\bigg)
\]
if $n_u(t) > 0$, and $0$ otherwise. The first factor grows with the
number of occurrences of $u$ in $t$, and the second decreases with the
number of texts that contain $u$. A term that occurs often in $t$ and
in few other texts thus weighs more in $t$ than a term that occurs
in most texts and rarely in $t$. The vector
$x(t) := (x_u(t))_{u \in V}$, divided by its norm, records which terms
$t$ uses and how characteristic of $t$ each of them is. The inner
product of two such normalized vectors, their cosine similarity, which we call cosine for short, is
meant to measure how similar the two texts are. It only counts the
terms that both texts use, so that two texts on the same subject that
have no term in common get orthogonal vectors, as if they were
unrelated. Latent semantic analysis \citep{deerwester1990lsa}
corrects this by projecting the
vectors on a small number $k$ of directions, chosen so that terms that
often occur in the same texts of the corpora are projected onto nearby vectors. Two texts on the same
subject then get nearby representations, even without a common term. Let $X$ be the matrix
whose rows are the normalized vectors $x(t)/\|x(t)\|$ of all the texts
of the corpora, and $P$ the matrix whose columns are its top $k$ right
singular vectors. These columns form an orthonormal basis of the
subspace of dimension $k$ closest to the rows of $X$, in the sense that
it minimizes the sum of the squared distances of the rows to it. We call \emph{word-statistics representation} of
$t$ the vector
\[
a^{\mathrm{ws}}(t) := P^{\top} \frac{x(t)}{\|x(t)\|} \in \mathbb{R}^{k}.
\]
The quality of the representation depends on $k$. On a synonym test,
for instance, latent semantic analysis does best around 300 directions, and much worse with very few
directions or with no projection at all \citep{landauer1997solution}.
We fix $k = 512$, of the same order, and do not tune it. Building concept vectors and scores
from $a^{\mathrm{ws}}(t)$ in the same way as from $a^{(\ell)}(t)$ yields
the \emph{word-statistics concept vector} $v_c^{\mathrm{ws}}$ and the
\emph{word-statistics score} $s_c^{\mathrm{ws}}(t)$. Every experiment
of the following sections that uses concept vectors is also done with
word-statistics concept vectors (\cref{sec:intro}). We also regress the
model's scores on the word-statistics scores, and recompute the
statistic on the residuals, which keeps only what the model's scores do
not share with word statistics. If a structure found in concept vectors
does not appear in word-statistics concept vectors, this is evidence
that the model captures something that classical word statistics do
not. Otherwise, the comparison cannot distinguish the model from word
statistics, although the model may represent the structure in a richer
way.

\subsection{Confidence intervals}
\label{app:ci}

Some results come with
95\% bootstrap intervals. A bootstrap interval estimates how much a
statistic would change with other data of the same kind. It draws many
resamples of the data, recomputes the statistic on each, and keeps the
central 95\% of the values. A resample draws as many units as the data
contain, at random and keeping repeats, so that it has the same size
as the data. For a statistic over pairs of structures, such as a
structure and one of its ancestors, the units are the structures: a
resample draws structures in this way, and the statistic is recomputed
on all the pairs between the drawn structures. We draw structures
because pairs that share a structure depend on the same texts, and in
another bank they would be present or absent together. Drawing pairs
one by one would make the resamples more alike than other banks would
be, and the intervals too narrow. This construction is called a
\emph{cluster bootstrap}.

\subsection{Steering and readouts}
\label{app:steer-methods}

We steer by activation
addition \citep{turner2023steering,rimsky2024steering}. A steering run
fixes a layer $\ell$, a steering unit vector $u$ and an intensity $\alpha$, and
lets the model continue a prompt, token by token, for at most 400
tokens. At each forward pass, right after layer $\ell$ is computed, we
add $\alpha\, \bar n_\ell\, u$ to the residual stream, and the
layers above $\ell$ compute from the modified states. No other layer is
modified. Here $\bar n_\ell$ is the median euclidian norm of the activations at
layer $\ell$, at the last token of reference texts, and measuring $\alpha$ in this unit
makes it comparable across layers and models. The first forward pass reads the prompt, and the vector is
added to the activation of each of its tokens at layer $\ell$. Each
following pass produces one token, and the vector is added to the
activation of this token at layer $\ell$. We call the generated text a
\emph{steered completion}. Its length is of the order of the corpus
texts (at most 200 words). Each steered completion is compared
with an \emph{unsteered completion}, generated from the same prompt
with the same random seed without steering, so that the two differ only by the
steering. The layer
$\ell$, the reference texts and the values of $\alpha$ depend on the
model and on the experiment, and each experiment description states
them. We measure the effect of steering in two ways. The
\emph{readout} measures whether the model's own representation of the
completion moves toward the concept. The completion is passed again
through the model, without the prompt and without steering, since the
activations produced during generation contain the added vector. Its
activations at layer $\ell$ are averaged, and this mean is projected on
every unit concept vector. For each concept, the difference between the
readouts of a steered completion and of its unsteered completion,
averaged over prompts, is divided by the standard deviation of the
unsteered readouts across prompts. Since the readout is computed with
the concept vectors themselves, it does not tell whether the completion
is more about the concept, which is what steering is meant to produce.
A \emph{judge} (Qwen2.5-72B-Instruct) measures this from outside the
model. It sees each steered completion and its unsteered completion in
random order, without knowing which is which, and says which is more
about the concept. Finally, we \emph{calibrate} both measurements on
real content. We insert 1, 2 or 4 real sentences about the concept into
unsteered completions, which shows whether the judge detects real
content, and how far real content moves the readout
(\cref{sec:steering}).

\subsection{Jacobian lens and writing vectors}
\label{app:writing}

\paragraph{Jacobian lens.}  The Jacobian lens
\citep{anthropic2026jlens} describes how a change of the state of the
residual stream at a layer propagates to the final states. It is
constructed as follows. Fix a text, a token $t$ of this text and a
token $t'$ that is $t$ or comes after it. Let $h_\ell \in \mathbb{R}^d$
be the state of the residual stream at token $t$ after layer $\ell$,
and $h_L \in \mathbb{R}^d$ the state at token $t'$ after the last layer
$L$. Replace $h_\ell$ by an arbitrary vector $x$, and let the layers
above $\ell$ compute as usual. The final state at token $t'$ becomes a
vector $F(x)$. This defines a function
$F : \mathbb{R}^d \to \mathbb{R}^d$ with $F(h_\ell) = h_L$. Its Jacobian
at $h_\ell$ is the $d \times d$ matrix of partial derivatives
$\partial F_i / \partial x_j$, taken at $x = h_\ell$. Adding a small
vector $\delta$ at layer $\ell$ at token $t$ changes the final state at
token $t'$ by about this matrix times $\delta$. The \emph{Jacobian lens}
of layer $\ell$ is the mean $J_\ell$ of these matrices over the texts
of a corpus and over all the pairs of tokens $t \le t'$ of each text. It
describes how a change at layer $\ell$ at one token propagates to the
final states of this token and of the following ones, on
typical text. We compute it for each model with the code released by
its authors, on 128-token excerpts of BBC news
articles,\footnote{Hugging Face dataset
\texttt{RealTimeData/}\allowbreak\texttt{bbc\_news\_alltime}.} which are generic texts
unrelated to mathematics.

\paragraph{Writing vectors.}  The \emph{writing
vector} of a token at layer $\ell$ is the direction that, added at layer
$\ell$, most raises the model's output score for this token. We obtain
it from the Jacobian lens as follows. After the last layer, the model
turns the final state $h_L$ at a token
into a score for each possible next token, its \emph{logit}. It first
normalizes $h_L$: it divides it by the root mean square of its
coordinates, $\operatorname{rms}(h_L)$, and multiplies it coordinate by
coordinate by a learned vector $g$. The logit of a token $v$ is then the
inner product of the result with a learned vector $W_U[v]$, the
\emph{output embedding} of $v$:
\[
\operatorname{logit}_v = \frac{\langle W_U[v], g \odot h_L\rangle}{\operatorname{rms}(h_L)} = \frac{\langle g \odot W_U[v], h_L\rangle}{\operatorname{rms}(h_L)}.
\]
A small change of $h_L$ barely changes $\operatorname{rms}(h_L)$. Adding
$\delta$ at layer $\ell$ at one token thus changes, on average, the
logit of $v$ at this token or a later one by about $\langle J_\ell^{\top} (g \odot W_U[v]), \delta \rangle /
\operatorname{rms}(h_L)$,
which, for a given norm of $\delta$, is largest when $\delta$ is along
$J_\ell^{\top} (g \odot W_U[v])$. Adding $\delta$ at every token, as in
steering, adds up such effects, so that the same direction
approximately remains the one that most raises the logit of $v$, for a
given norm of $\delta$. We call
the direction of $J_\ell^{\top} (g \odot W_U[v])$ the writing vector of $v$ at layer
$\ell$. The writing vector is thus defined for a single token, as in
\citet{anthropic2026jlens}. For a concept whose name has only one token,
the writing vector of this concept is the writing vector of this token.
For a concept whose name has more than one token, such as \emph{metric
space}, we use the writing vector of the first token of its name. It
cannot be interpreted directly as the writing vector of the concept
itself, only as a proxy. We report the results on concepts
whose name has more than one token separately.

\subsection{Construction of the concept vectors}
\label{app:variants}

\Cref{sec:method} builds the vector of a concept by subtracting the
mean over all concept corpora (centering). The most common construction
in the steering literature subtracts a single baseline corpus instead.
Here we compare five constructions and show why we center. The five
constructions are concept minus a non-mathematical corpus (history,
biology, cooking, law, music); concept minus a corpus of generic
mathematics; the centered vector; concept minus generic mathematics
with the mathematics direction projected out, where the mathematics
direction is the generic-mathematics corpus minus the
non-mathematical one; and concept minus generic mathematics with the
top 10 principal components (PCs) of the pooled texts removed.

Vectors built against a single baseline share a large common
component. The mathematics direction is 1.4--1.6 times as long as the
median concept-minus-generic-mathematics vector, and the
concept-minus-non-mathematics vectors have a mean cosine of 0.71--0.74
with it. Two unrelated concepts therefore have nearly parallel vectors,
and a child text scores almost as high on an unrelated structure as on
its parent. \Cref{tab:variants} shows this on all four models: with a
single baseline, the mean cosine between concept vectors is 0.68--0.84
and the parent and unrelated ratios differ by 0.04--0.11. Centering
removes the common component by construction (mean cosine 0.00),
brings the unrelated ratio to zero, and gives the best separation AUC and
the best separation of parents from unrelated structures on every
model. Removing principal components helps, but less, and leaves the
vectors aligned.

\begin{table*}[t]
\centering
\small
\caption{Five constructions of the concept vectors (layer 16, 136
concepts, separation AUC). ``cos'': mean cosine between two
concepts' vectors. ``Parent'' and ``unrel.'': ratio $r$ of a child's
texts on its direct parent and on an unrelated structure. ``AUC'':
parent against unrelated pairs.}
\label{tab:variants}
\begin{tabular}{llcccc}
\toprule
Model & Vectors & Separation & cos & Parent / unrel. & AUC \\
\midrule
R1-Qwen-7B & $-$ non-math & 0.60 & 0.80 & 0.98 / 0.91 & 0.63 \\
 & $-$ generic math & 0.67 & 0.73 & 0.96 / 0.86 & 0.78 \\
 & math direction removed & 0.72 & 0.63 & 0.94 / 0.79 & 0.82 \\
 & top 10 PCs removed & 0.74 & 0.78 & 0.97 / 0.91 & 0.80 \\
 & centered & \textbf{0.75} & 0.00 & 0.76 / $-$0.03 & \textbf{0.90} \\
\midrule
R1-Qwen-32B & $-$ non-math & 0.58 & 0.82 & 0.98 / 0.93 & 0.62 \\
 & $-$ generic math & 0.63 & 0.75 & 0.96 / 0.90 & 0.62 \\
 & math direction removed & 0.71 & 0.63 & 0.93 / 0.81 & 0.76 \\
 & top 10 PCs removed & 0.74 & 0.75 & 0.97 / 0.89 & 0.83 \\
 & centered & \textbf{0.74} & $-$0.01 & 0.81 / $-$0.06 & \textbf{0.91} \\
\midrule
R1-Llama-8B & $-$ generic math & 0.66 & 0.68 & 0.95 / 0.85 & 0.67 \\
 & centered & \textbf{0.77} & 0.00 & 0.73 / 0.00 & \textbf{0.90} \\
\midrule
Qwen2.5-Math-7B & $-$ generic math & 0.65 & 0.79 & 0.97 / 0.91 & 0.73 \\
 & centered & \textbf{0.73} & 0.00 & 0.76 / $-$0.02 & \textbf{0.90} \\
\bottomrule
\end{tabular}
\end{table*}

Parents separate from unrelated structures here, at AUC 0.90. This
does not contradict \cref{sec:inclusion}: an unrelated structure is far
from the child in $G$, and the ratio falls with distance in $G$.
The containment AUC of \cref{sec:inclusion} compares ancestors with
unrelated pairs at the same distance, and that comparison is at chance.

\section{Steering}
\label{app:steering}

This appendix details the steering experiments of \cref{sec:writing}.
Steering, the readout and the judge are defined in \cref{sec:method}
and \cref{app:steer-methods}. The readout of a concept is the
projection on its concept vector of the activations of a completion
when it is read again by the model without steering, and effects on
it are in standard deviations of the unsteered readout. The score
increase is the negative log-likelihood per token of a steered
completion under the unsteered model, minus that of its unsteered
completion, averaged over prompts. We steer twelve parent--child pairs,
by the concept vector of the parent, on texts about the child concept,
and the concept vectors of 53 concepts on non-conceptual prompts.

\subsection{Regimes of steering}
\label{app:regimes}

\Cref{fig:steering_regimes} classifies every intensity of the sweep,
for each model, type of text and sign, into four regimes. No effect:
the readout of the parent moves by less than 0.1 standard deviations
of the unsteered readout. Readout follows the steering direction: it
moves by at least 0.1 in the steering direction, and the score
increase is less than 0.2. Text degraded: the same, with a score
increase of 0.2 or more. Readout erratic: it goes against the steering
direction by more than 1, or falls below half of its largest value at
a smaller intensity. Boundaries are placed halfway, on a logarithmic
scale, between the tested intensities. These thresholds are our
choice. The results behind the figure are as follows.

\emph{Small intensities.} On texts about the child concepts, the
readout of the parent stays at its unsteered value up to
$\alpha \approx 0.10$--$0.15$ on R1-Qwen-32B and 0.24--0.36 on
R1-Qwen-7B. Up to $\alpha = 0.24$, the score increase is at most
0.07 on R1-Qwen-7B and 0.04 on R1-Qwen-32B, for both adding and
subtracting the parent.

\emph{The window.} Among the tested intensities, it goes from
$\alpha = 0.15$ to 0.55 on R1-Qwen-32B, and from 0.24 (subtracting) or
0.36 (adding) to 0.55 on R1-Qwen-7B. Over the window, the score
increase grows up to 0.24--0.26 on R1-Qwen-32B and 0.45--0.55 on
R1-Qwen-7B. Centered vectors move their concept on
non-conceptual prompts ($+0.35$ standard deviations on
R1-Qwen-32B at $\alpha = 0.36$), and on these prompts, at
$\alpha = 0.3$, the concept vector moves its own concept in the
steering direction for 48 of the 53 concepts on R1-Qwen-32B (36 on
R1-Qwen-7B). On texts about the
child concept, the parent moves more than its children, siblings and
ancestors on average, and much more than the other concepts, but its
median rank among the 136 concepts is only 17th to 48th. The
text changes much less than the readout. On non-conceptual prompts,
the frequency of the name of the steered concept stays at most 0.01
occurrences per 100 words on R1-Qwen-7B, with or without steering. The calibration shows that the judge is sensitive where it
matters. On non-conceptual prompts, one inserted sentence from
the parent's corpus that never names the parent moves the readout by
$+0.15$ standard deviations and the judge's verdict by $+0.33$, while
unrelated sentences, which disrupt the text as much, give $0.00$.
Steering moves the readout further ($+0.21$ at $\alpha = 0.55$), and
yet the judge does not see it ($+0.00$ to $+0.01$, with ``same'' in
95--98\% of comparisons). On texts about the child concept, the
frequency of the name of the parent does change. We measured it on a
separate set of generations of R1-Qwen-7B, made at $\alpha = 0.1$,
0.36 and 0.55, of which only the last two are inside the window. At
these two intensities, adding the parent raises it from 0.51 to 0.69
and 0.73 occurrences per 100 words, and subtracting the parent lowers
it to 0.43 and 0.33. The judge sees the removal: subtracting the
parent makes the text less about it ($-0.18 \pm 0.04$ at
$\alpha = -0.36$, against $-0.11$ for random directions). It does
not see the addition: over the window of R1-Qwen-7B, its verdict stays
between $-0.12$ and $+0.07$, with 95\% intervals of about $\pm 0.15$
that contain 0, on about 70 comparisons per intensity.

\emph{Large intensities.} At $\alpha = 0.85$, the score
increase is 0.89--1.38 on R1-Qwen-7B and 0.74--0.89 on R1-Qwen-32B. From
$\alpha = 1.3$ on, the readout of the parent no longer moves in the
steering direction. If the concept vector wrote its concept,
subtracting the parent would lower this readout. Instead, it raises it
to between $+28$ and $+209$ standard deviations of the
unsteered readout. This rise is not specific to the parent. At
$\alpha = -1.3$, the readouts of all concepts rise: those of its
siblings, children and ancestors by $+25$ to $+143$, and those of the
other concepts by $+0.8$ to $+9.5$ on average, and the parent ranks
only 62nd to 68th of the 136 concepts. This rise of the readout of the
parent thus does not mean that the text has become more about it,
which subtracting the parent would not be expected to produce. Adding
the parent, which would then raise its readout, moves this readout by
between $-12$ and $+3$ standard deviations, and at $\alpha = 2$ the
readout of the parent is below its unsteered value on both models. The
completions we inspected at these intensities are incoherent: repeated
connectives when adding, stray punctuation when subtracting. At high
intensity, vectors built against a single corpus of generic mathematics
(uncentered vectors) write a register of derivation. At $\alpha = 1.3$,
94\% of the words are connectives (``So Therefore Then Let \dots''),
which neither random directions nor the mathematics direction produce.

\subsection{Intensity sweep, judge and variations of the intervention}
\label{app:sweep}

\Cref{sec:steering} found that steering along a concept vector moves the
readout as real content would, without putting the content in the
text. This section gives the intensity sweep and the variations of
the intervention that do not change that conclusion. Effects are in
standard deviations of the unsteered readout. Some variations were run
with the uncentered vectors (concept minus generic mathematics); we say
so where it applies.

\paragraph{Intensity sweep.} We steer twelve parent--child pairs at 15
log-spaced intensities from 0.005 to 2, with both signs, on texts about the child concept, non-conceptual
and unrelated prompts, and steer each of 53 concepts on
non-conceptual prompts. \Cref{tab:sweep}
gives the effect of the parent at $\alpha = 0.36$, with the effect on
the child in parentheses.

\begin{table*}[t]
\centering
\small
\begin{tabular}{lcccc}
\toprule
 & \multicolumn{2}{c}{R1-Qwen-7B} & \multicolumn{2}{c}{R1-Qwen-32B} \\
 & uncentered & centered & uncentered & centered \\
\midrule
child texts, $-$ & $-0.37$ ($-0.47$) & $-0.20$ ($-0.16$) & $-0.32$ ($-0.37$) & $-0.29$ ($-0.22$) \\
child texts, $+$ & $+0.28$ ($+0.21$) & $+0.05$ ($-0.07$) & $+0.27$ ($+0.23$) & $+0.20$ ($+0.05$) \\
non-conceptual prompts, $+$ & $0.00$ & $+0.13$ & $+0.25$ & $+0.35$ \\
53 concepts & $+0.02$ (55\%) & $+0.08$ (68\%) & $+0.13$ (79\%) & $+0.19$ (91\%) \\
onset $\alpha^*$ ($+$ / $-$) & 0.15 / 0.15 & 0.36 / 0.24 & 0.10 / 0.15 & 0.10 / 0.15 \\
\bottomrule
\end{tabular}
\caption{Intensity sweep (\cref{app:steering}): effect on the parent at $\alpha = 0.36$, in standard deviations of the unsteered readout, with the effect on the child in parentheses. Rows marked $-$ and $+$ subtract and add the parent's vector. 53 concepts: each concept steered on non-conceptual prompts, effect and share of concepts that move in the steering direction. Onset: intensity at which the effect starts, estimated by fitting,
on the intensities below 0.7, a curve that is constant up to the onset
and linear in $\log \alpha$ above it.}
\label{tab:sweep}
\end{table*}

Centered vectors steer on
non-conceptual prompts, where the uncentered ones do little, and on texts about the child
concept they move the parent without dragging the child. Specificity
stays modest: at $\alpha = 0.36$ and 0.55, the median rank of the
parent among the 136 concepts is 17th to 45th. Text quality, measured by the negative log-likelihood of the
completion, holds up to $\alpha = 0.55$ on R1-Qwen-32B and collapses
above 0.85.

\paragraph{What the judge sees.} Over 35{,}600 comparisons on the
uncentered sweep, steered texts are never rated more about the target
(within $\pm 0.06$ up to $\alpha = 0.36$), and on non-conceptual prompts the
parent never appears (``same'' in 99--100\% of comparisons). Coherence
drops from $\alpha \approx 0.36$, following the log-likelihood. In the
judge's free-text descriptions, the only consistent theme is register
(``more conversational, less structured''), which rises from about 5\%
to 11--23\% at $\alpha = 0.55$ for both signs. A 10\% sample of the
centered sweep gives the same null.

\paragraph{Calibration of the judge.} Into unsteered completions we
insert $k = 1$, 2 or 4 sentences, at random sentence boundaries, from the
parent's corpus (from texts that name the parent, or that never do) or
from an unrelated concept's corpus, and the judge compares each edited
text with its original. The net score is the probability that the
edited text is judged more about the parent minus the probability that
the original is. \Cref{tab:calibration} gives the results.

\begin{table*}[t]
\centering
\small
\begin{tabular}{lcc}
\toprule
Inserted sentences & Non-conceptual, $k$ = 1 / 2 / 4 & Child, $k$ = 1 / 2 / 4 \\
\midrule
parent, named & $+0.98$ / $+0.99$ / $+1.00$ & $+0.44$ / $+0.43$ / $+0.25$ \\
parent, unnamed & $+0.33$ / $+0.45$ / $+0.63$ & $-0.24$ / $-0.49$ / $-0.66$ \\
unrelated concept & $+0.01$ / $-0.02$ / $-0.06$ & $-0.27$ / $-0.56$ / $-0.70$ \\
readout, unnamed parent & $+0.15$ / $+0.29$ / $+0.57$ & $-0.01$ / $+0.05$ / $+0.18$ \\
\bottomrule
\end{tabular}
\caption{Calibration of the judge (\cref{app:steering}): net score for $k$ inserted sentences, and readout shift for $k$ unnamed parent sentences.}
\label{tab:calibration}
\end{table*}

On non-conceptual prompts the judge detects one unnamed parent
sentence, which moves the readout by $+0.15$, and ignores unrelated
sentences that disrupt the text as much; it reads content and not
damage. Steering that moves the readout by $+0.21$ is not detected. On
texts about the child concept the judge reacts only to names, since unnamed parent
sentences score like unrelated ones, so judge nulls on texts about the child concept
say nothing unless names are involved. Unrelated sentences lower the
parent's readout ($-0.15$ to $-0.33$), so the readout does not simply
measure ``more mathematics''. Even with real content inserted, the
parent ranks only 44--86 of 136, so part of the poor specificity of
steering is a property of the readout itself.

\paragraph{Variations of the intervention.} None of the following makes
the text change in a concept-specific way, as judged.
\emph{Neighbourhood} (uncentered): directions at 5$^\circ$ to 90$^\circ$
from the concept vector keep about 90\% of the readout effect at
30$^\circ$ and half of it only at 45--60$^\circ$; interpolating between
two concept vectors gives neither a switch nor a blend.
\emph{Other directions} (uncentered): random directions and principal
components move the readout profile about as much as concept vectors
(mean absolute displacement 0.1--0.2). One offset works in
part: the offset from monoid to group at $\alpha = 0.3$ on monoid prompts
raises group by $+0.42$ and lowers monoid by $-0.59$; other offsets that
add inverses are weak, and the commutativity and identity offsets do
nothing. \emph{Clamping} the projection to a target level instead of
adding: setting it to zero suppresses more cleanly than subtracting
($-0.58$ at a log-likelihood cost of $+0.11$); clamping to 1--1.5
times the concept's natural level does not amplify, and from 2 times
the text degrades. \emph{Steering layer}: layers 8--16 of R1-Qwen-7B work
best, layers 4 and 24 about half as well, and spreading the budget
over layers 8--16 gives $-0.39$ / $+0.24$ at $\alpha = 0.3$ at the
same cost.

\paragraph{What high intensity writes.} At $\alpha = 1.3$ the
uncentered ring vector, added while R1-Qwen-7B explains why every nonzero
quaternion has an inverse, produces ``So The Multip Let So Let m So Then
Multiplying Both S Multiplying Therefore So Similarly Therefore\dots''.
A random direction collapses differently (``reciprocal [rec] in
reciprocal [blue] * [blue] * * *''). Read through the Jacobian lens,
the component shared by the uncentered vectors writes connectives
(`` Moreover'', `` Furthermore'' at layer 8 of R1-Qwen-7B; `` So'',
`` Thus'', `` Hence'' near its output; `` Therefore'', `` Thus'' and
the Chinese words for ``therefore'' and ``however'' at layers 56--60 of
R1-Qwen-32B). The uncentered vectors therefore push a register of
mathematical derivation, which is the component that centering
removes.

\subsection{Reading and writing vectors}
\label{app:readwrite}

This section completes \cref{sec:readwrite}, which compares, for 48
emotions and 136 mathematical concepts, the concept vector of a
concept with its writing vector, the direction that most raises the
logit of the first token of its name (\cref{app:writing}).

\paragraph{Groups.} The three groups are 48 emotions, 35
mathematical concepts named by a common word (ring, field, average,
\dots), whose names have other, non-mathematical meanings, and 101
other mathematical concepts, and 43 and 15 when restricted to
single-token names. For names of more than one token, we use the
writing vector of the first token, which is only a proxy. Drawing sets of 12 in
every group gives them all the same chance level for the
identification rate. The mean cosine between a concept vector and its
writing vector is $+0.03$ and $+0.02$ on R1-Qwen-7B and R1-Qwen-32B for
the mathematical concepts named by a common word, and $+0.02$ and
$+0.01$ for the others, and the cosine is positive for 78\% and 60\% of
the mathematical concepts. Single-token mathematical names are
identified somewhat better than the others on R1-Qwen-32B (0.28
against 0.19 at layer 36), and no better on R1-Qwen-7B.

\paragraph{Steering.} We compare how much the concept vector, the
writing vector and a random direction raise the probability of
writing the name, for emotions and mathematical concepts. This differs
from the steering of \cref{sec:steering} in three ways. There, we
added the concept vector of a parent while the model generated, and
measured the content of the generated text, for mathematical concepts
only. Here, the texts are fixed and the same for all groups: unsteered
completions of the same prompts, which we pass through the model with
a vector added. At each position, we compare the probabilities that
the model gives to the next token with and without the vector, and we
rank all tokens by how much the vector raises their log-probability,
on average over the positions. We report the rank of the first token
of the concept's name. Finally, the writing vector serves as a
reference, since it is built to write the name. We steer at layer 16
of R1-Qwen-7B and layer 36 of R1-Qwen-32B, with the intensity measured
relative to the typical norm of the activations at this layer on the
mathematical corpora (\cref{sec:method}).

Steering with the writing vector puts the name at median rank 1 in
every group, on both models, and random directions put it at median
ranks between 3{,}500 and 106{,}000. Steering with the concept vector
puts it at median rank 160--200 for mathematical concepts with a
common name and 1{,}500--3{,}800 for the others on R1-Qwen-7B, and
1{,}900--3{,}700 and 16{,}000--29{,}000 on R1-Qwen-32B
(\cref{tab:readwrite}), the ranges covering $\alpha = 0.36$ and 0.55.
For emotions, it puts the name at median rank 46--60 and 20--40, and
into the generated text. On texts about the child concept, for the
twelve parent--child pairs of \cref{sec:steering}, the concept vector
of the parent puts its name at median rank 236 and 182 at
$\alpha = 0.55$, and the writing vector at rank 1 in 23 of the 24
cases. Across the 48 mathematical concepts, the rank of the name
follows the cosine of the concept vector with its writing vector: the
Spearman correlation between the cosine and the logarithm of the rank,
at $\alpha = 0.55$, is $-0.75$ on R1-Qwen-7B and $-0.85$ on
R1-Qwen-32B. The ten concepts with the largest cosines, which are
comparable to those of emotions, put their name at median rank 13 and
131, and the ten with the smallest cosines at median rank 1{,}340 and
88{,}000.

\section{Structure of concept vectors}
\label{app:structure}

This appendix details the experiments of \cref{sec:geometry}, on the
four models of \cref{sec:method}: separation of concepts, the
geometry of concept vectors across layers, inclusion and composition.
The score $s_c(t)$ of a text $t$ on a concept $c$ is defined in
\cref{app:scores}, and the word-statistics score $s_c^{\mathrm{ws}}(t)$
in \cref{app:wordstats}.

\subsection{Separation}
\label{app:separation}

\paragraph{Measures.} We call \emph{separation AUC} of a concept $c$ at layer $\ell$ the
AUC of the score $s_c^{(\ell)}$ between the texts of $C_c$ and all
other texts. We compute it at every layer and average it over concepts,
on four models. The score uses the activation of the last token of a
text. We also compute the separation AUC with the mean activation over
the tokens of the text instead of the activation of its last token. We
compare the layers with two other representations of the texts, which
we call \emph{references}. The first is the input embeddings, taken in
the same way as the activations (the embedding of the last token, or
the mean of the embeddings over the tokens), with the score computed as
for a layer. The second is word statistics, with the score
$s_c^{\mathrm{ws}}$.

\paragraph{Results.} Unless a model is named, a range of values is
taken over the four models, and over the layers of the plateau when the
plateau is mentioned. With the last-token activation, the separation
AUC is 0.49--0.50 on the input embeddings, already
0.65--0.67 after the first layer, and rises over about the first
quarter of the depth. It then stays on a plateau up to the
second-to-last layer (0.74--0.78 on R1-Qwen-7B, 0.73--0.78 on
R1-Qwen-32B, 0.77--0.79 on R1-Llama-8B, 0.72--0.77 on Qwen2.5-Math-7B),
and drops at the last layer (0.66--0.74). On the plateau, the separation AUC varies by at
most 0.05 from one layer to another. The layer where it is highest is
not stable. We recompute it on 30 random halves of the texts of every
concept: on R1-Qwen-32B it is layer 24 for 10 halves and layer 60 for
the other 20, and it also moves across the plateau on R1-Qwen-7B and
R1-Llama-8B, while it stays at layers 23--25 on Qwen2.5-Math-7B. With
the mean activation, the separation AUC is 0.95--0.97 on the plateau,
and already 0.90--0.91 on the input embeddings. With word statistics, it
is 0.98 on the generated corpora and 0.99 on the Wikipedia corpora,
above every layer of every model (on the Wikipedia corpora, R1-Qwen-7B
reaches 0.82--0.84 at layers 12--20 with the last-token activation).

The separation AUC of a layer measures how it separates concepts
over all texts. To have more precise information on how each layer
separates concepts, we look at texts that a reference (for instance
word statistics) scores approximately equally. What the layer separates
among these texts, the reference cannot separate. For a concept $c$, we
split all the texts into 20 bins of equal size by their reference score
on $c$. In each bin, we compute the AUC of the score $s_c^{(\ell)}$
between the texts of $C_c$ and the other texts, and we average these
AUCs over the bins, weighting each bin by its number of pairs made of a
text of $C_c$ and another text. We call the result the
\emph{conditional separation AUC} of $c$ at layer $\ell$ given the
reference, and average it over concepts. With the input embeddings as the reference, the conditional
separation AUC of the layers of the plateau is 0.75--0.84 with the
last-token activation and 0.86--0.91 with the mean activation. Computed
in the same bins, the conditional separation AUC of the input
embeddings themselves is 0.11--0.13 and 0.66 respectively. For a concept $c$, the score $s_c^{(\ell)}$ of a layer
on the plateau and the word-statistics score $s_c^{\mathrm{ws}}$ rank
the texts differently: their Spearman correlation over all texts,
averaged over concepts, is 0.19--0.27 with the last-token activation
and 0.55--0.67 with the mean activation.
Yet with word statistics as the reference, the conditional
separation AUC of the layers of the plateau is only 0.59--0.64 with the
last-token activation and 0.76--0.80 with the mean activation, while,
computed in the same bins, that of word statistics themselves is
0.86. The layers thus rank the texts differently from word
statistics, but separate concepts less well than word statistics, both
over all texts and within the bins.

The results also suggest how a text could be read better. The mean
activation separates concepts better than the last-token activation,
so the information is spread over the tokens of the text. A large part
of this advantage is lexical, since the mean of the input embeddings
already separates concepts at 0.90--0.91, but not all of it, since the
mean activation also separates concepts better given word statistics.
Subtracting the mean of the input embeddings from the mean activation
would remove the purely lexical part.

\subsection{Layers}
\label{app:layers}

\Cref{sec:storage} found that concepts are separable on a broad
plateau of layers and that no single layer stands out. Here we follow
the concept vectors across layers, to see whether a concept keeps its
direction through the network. We compare each concept with 20
pseudo-concepts, each built from 90 texts drawn at random from the
pooled corpora, which give the behavior of a vector that carries no
concept.

A single concept's vector drifts. The cosine between its vectors at two
adjacent layers is 0.88--0.90, it decays over about 10--11 layers, and
the pseudo-concepts give the same numbers, so this drift is a property
of the residual stream and not of the concepts. The relations between
concepts are stable. The representational similarity (RSA) between the
matrices of pairwise similarities of all concepts at two layers is
0.98--0.99, far above any single concept's cross-layer cosine. Much of
the drift is one shared transformation. A linear map between two
layers, orthogonal on the span of the concept vectors, fitted on 75\%
of the concepts, predicts the vectors of the held-out 25\% better than
the identity on all four models (held-out cosine 0.94--0.98 instead of
0.87--0.96 for adjacent layers, 0.79--0.91 instead of 0.58--0.79 four
layers apart). The cosines between layers fall into three blocks, with
boundaries at about 25\% and 70\% of the depth on all four models.

Two caveats apply to these numbers. First, the same texts are used at every
layer, so their sampling noise is shared across layers and can inflate
both the RSA and the fitted map. Second, the split-half reliability of a
centered concept vector is about 0.70 in the middle layers, which is
enough for the population statistics of the paper but caps the cosine
that any single concept can show.

The separation AUC of R1-Qwen-7B by layer is given in \cref{tab:layers}.

\begin{table*}[t]
\centering
\small
\begin{tabular}{lccccccccc}
\toprule
Layer & emb. & 1 & 4 & 8 & 12 & 16 & 20 & 24 & 27 \\
\midrule
Last-token activation & 0.49 & 0.66 & 0.73 & 0.73 & 0.77 & 0.75 & 0.75 & 0.77 & 0.71 \\
Mean activation & 0.91 & 0.92 & 0.93 & 0.95 & 0.96 & 0.96 & 0.95 & 0.96 & 0.90 \\
\bottomrule
\end{tabular}
\caption{Separation AUC of R1-Qwen-7B by layer (\cref{app:layers}), centered vectors.}
\label{tab:layers}
\end{table*}

\subsection{Inclusion}
\label{app:inclusion}

\paragraph{Definitions.} For two concepts $c$ and $c'$, we
define the \emph{ratio}
\[
r(c\to c') := \frac{\langle v_c, v_{c'} \rangle}{\frac{1}{|C_{c'}|}
\sum_{t\in C_{c'}} \langle a(t)-\bar\mu, v_{c'}^{-t} \rangle}.
\]
Since $v_c$ is the mean of $a(t)-\bar\mu$ over the texts of $C_c$, the
numerator is the mean of $\langle a(t)-\bar\mu, v_{c'} \rangle$ over
the texts of $c$, and the denominator is the same mean over the texts
of $c'$, with $v_{c'}^{-t}$ in place of $v_{c'}$. The ratio thus
measures how much the texts of $c$ score on $c'$ compared with the
texts of $c'$, both measured from the reference $\bar\mu$. With
$v_{c'}$ in place of $v_{c'}^{-t}$, the denominator would be
$\|v_{c'}\|^2$, and the ratio would be the coefficient of the
projection of $v_c$ on $v_{c'}$. We use $v_{c'}^{-t}$ so that no text of
$c'$ is scored on a vector it contributed to, in the same way as in the
definition of the score (\cref{app:scores}). We call \emph{ancestor pair} a pair $(c, c')$ where
$c'$ is an ancestor of $c$.

\paragraph{Containment and directionality.} We measure two properties of inclusion. (i) \emph{Containment}:
if concept vectors respect inclusion, the texts of a concept should
score higher on the concept vectors of its ancestors than on those of
other concepts. The ratio may depend on the distance in $G$
alone, and ancestor pairs are on average at a smaller distance than
unrelated pairs. To separate inclusion from distance, we compare the
ratio of an ancestor pair only with the ratios of unrelated pairs at
the same distance. For a distance $k$, we take the
ancestor pairs $(c, c')$ where $c'$ is at $k$ steps from $c$ along the
edges of $G$, and the unrelated pairs at distance $k$. The \emph{containment AUC} is the share of
comparisons, between an ancestor pair and an unrelated pair at the
same distance, in which the ancestor pair has the higher ratio, over all
distances $k \ge 3$. Following \citet{park2025geometry}, we say that an
ancestor pair $(c, c')$ is \emph{hierarchically orthogonal} when $v_c -
v_{c'}$ is orthogonal to $v_{c'}$, which amounts to $r(c\to c') = 1$.
Containment is weaker: if all ancestor pairs were hierarchically
orthogonal, ancestor pairs would score as high as the texts of the
ancestor itself, above unrelated pairs.

(ii) \emph{Directionality}: inclusion is one-way. If $c'$ is an
ancestor of $c$, every instance of $c$ is an instance of $c'$, but not
conversely. A text about $c$ is thus also about $c'$, while a text about
$c'$ is in general not about $c$. If concept vectors respect inclusion,
the texts of $c$ should therefore score higher on $c'$, compared with
the texts of $c'$, than the texts of $c'$ score on $c$, compared with
the texts of $c$, that is, $r(c\to c') > r(c'\to c)$. We call
\emph{directionality} the share of ancestor pairs for which this
holds.

Both measures use 97 structures and 520 ancestor
pairs, on the four models at layers 12, 16 and 20. Since separation and
the geometry of the concept vectors vary little across the plateau
(\cref{sec:storage}), any layers of the plateau can be used. These three
layers lie in the plateau, except layer 12 of R1-Qwen-32B, which gives
the same results. We also run them
on the ratios computed from word statistics in place of the activations.
To keep only the part of the ratios of a model that word
statistics do not explain, we regress them linearly on the word-statistics ratios, over all pairs of
structures in the same connected component of $G$, and compute both measures on the residuals.

\paragraph{Results.} The ratio falls steadily with the distance in $G$, from a
concept to its ancestors and between unrelated structures in both orders (R1-Qwen-7B, layer 16: 0.42 for ancestor pairs
against 0.37 for unrelated pairs at distance 3, 0.34 against 0.22 at
distance 4, 0.26 against 0.18 at distance 5). On our
concept vectors, hierarchical orthogonality would give a ratio of 1 for
every ancestor pair. The mean ratio is 0.72--0.85 from a concept to its
parent, and 0.52--0.59 to an ancestor two steps away. \Cref{tab:containment}
gives both measures at layer 16. The containment AUC is 0.49--0.55 on every
model and layer, and the directionality is 0.54--0.59; every interval includes
chance. Word statistics give the same decay and the same null results
(containment AUC 0.50, directionality 0.47). The residuals no
longer decrease with the distance: their mean stays between $-0.13$ and
$0.06$ at every distance, except for R1-Qwen-32B at layer 12, where the
regression explains almost nothing. On the residuals, the containment AUC is 0.50--0.56
and directionality 0.53--0.58, so removing what word statistics explain
changes no conclusion. Two further analyses agree. On a subset of 32 structures and 120
ancestor pairs, dropping the texts of a concept that name one of its
ancestors leaves the ratios of ancestor pairs almost unchanged: their
mean is multiplied by 1.01--1.05. On the Wikipedia corpora, with 26
concepts and 95 ancestor pairs, the containment AUC is 0.50--0.54 and
directionality 0.49--0.60. The estimator of \citet{park2025geometry}, which whitens the
activations and is built to recover hierarchical orthogonality, does
not recover it on our concepts either (\cref{app:park}).

\paragraph{Estimator of \citet{park2025geometry}.}
\label{app:park}
 \citet{park2025geometry} estimate the vector of a concept in another
way, and we examine hierarchical orthogonality (\cref{sec:inclusion}) on
their vectors as well. They
first whiten the activations: they center them and transform them
linearly so that their covariance over all texts is the identity, which
keeps the directions of large variance from dominating. The vector
$\ell_c$ of a concept $c$ is then a linear discriminant: the direction
along which the whitened activations of the texts of $c$ lie farthest
from the mean of all texts, relative to their own spread. On a subset of 32
structures and 120 ancestor pairs, with R1-Qwen-7B and R1-Qwen-32B,
$\cos(\ell_{c'}, \ell_c - \ell_{c'})$ is $-0.67$ to $-0.68$ for a
concept $c$ and its parent $c'$, against $-0.70$ when $c'$ is a random
concept unrelated to $c$. Hierarchical orthogonality requires 0. For
two vectors of equal norm with cosine $\rho$, this cosine is
$-\sqrt{(1-\rho)/2}$, which is $-0.71$ for orthogonal vectors: the
vectors of a concept and of its parent are nearly orthogonal, like
those of unrelated concepts. The ratios computed from these vectors
give the same chance-level containment AUC (0.48--0.57).

\subsection{Composition}
\label{app:composition}

\paragraph{Definitions.} We call \emph{ingredients} of a
structure the concepts of the bank that its definition is composed of:
for instance, a monoid is a semigroup with an identity element, so its
ingredients are semigroup and identity element. A definition can also
contain parts that are not concepts of the bank: a Banach space is a
complete normed space, and completeness is not a concept of the bank,
so the only ingredient of a Banach space is normed space. We call
\emph{offset} from a concept $c$ to a concept $c'$ the difference $v_{c'}
- v_c$.

\paragraph{Measures.} (i) Composition: we fit the vector of each of the 14 structures for
which the bank lists ingredients (\cref{tab:ingredients}) by the closest
linear combination of the vectors of its ingredients (least squares),
and measure how much of the vector the fit explains by $R^2 = 1 -
\|v - u\|^2 / \|v\|^2$, where $v$ is the vector of the structure and
$u$ the closest combination: $R^2$ is 1 when the fit is exact and 0
when it explains nothing. With only
two or three vectors in thousands of dimensions, the closest
combination is far from the vector unless the two are related. If the vector of a structure is related to those of its
ingredients, its fit on its own ingredients should thus be much closer
than its fit on the ingredient list of another of these structures, or
on random concepts of the bank in the same number. We compare the three
fits.

(ii) Added requirements: we take 21 pairs of structures $(c, c')$
of the bank whose definitions differ by one added requirement, such as
having inverses (a group is a monoid where every element has an
inverse; \cref{tab:steps}). If adding the same requirement always moved
the vector by the same offset, the offsets of the pairs that add this
requirement would point in nearly the same direction, while offsets of pairs that add different
requirements would not. We measure how concentrated their directions
are: the mean cosine between the offsets of pairs that add the same
requirement, against the mean cosine between the offsets of pairs that
add different requirements.

(iii) Relations: the constructions of the bank apply one of four
relations (sub-object, morphism, quotient, product) to one of six
structures (\cref{sec:method}), and fill 22 of the 24 cells of this
grid. If applying the same relation always moved the
vector by the same offset, the offset from a structure to its
construction under a relation, such as from group to quotient group,
would be nearly the same for all six structures. We measure
the mean cosine between offsets of the same relation, against offsets
of different relations.

\paragraph{Results.} (i) At the middle layers, the fit of the vector
of a structure on the vectors of its ingredients explains $R^2 \approx
0.2$, against 0.12--0.17 for random concepts of the bank in the same
number, and the ingredients of the structure give a closer fit than the
ingredient list of another structure in about 60\% of cases. (ii) The
mean cosine between the offsets of pairs that add the same requirement
is 0.03--0.06, against 0.005--0.014 for pairs that add different
requirements (ranges over four models and three layers each). Word
statistics align the offsets of pairs that add the same requirement
more (mean cosine 0.17). (iii) The offsets of the same relation across
structures are barely aligned (mean cosine 0.06, against 0.02 between
offsets of different relations).

\paragraph{Interpretation.} The vector of a structure is barely closer
to a linear combination of the vectors of its ingredients than to a
combination of the vectors of other concepts, and adding the same
requirement, or applying the same relation, moves the vectors in
directions that are only weakly aligned. These measures do not exclude
a nonlinear composition rule, which the linear operations of
macro-interpretability could not use.

\label{app:definitions}

\Cref{tab:ingredients} gives the ingredients of the 14 structures of
the composition experiment (\cref{sec:composition}), and \cref{tab:steps} the
21 pairs of structures of the experiment on added requirements, grouped by
the requirement they add. Four of these six requirements (compactness, completeness,
finiteness and Hausdorff separation) are not concepts of the bank.

\begin{table}[h]
\centering
\small
\caption{Ingredients of the 14 structures of the composition experiment
(\cref{sec:composition}). The lists give the main parts of each
definition; some axioms, such as the symmetry of a distance, are not
concepts of the bank and are not listed.}
\label{tab:ingredients}
\setlength{\tabcolsep}{3pt}
\begin{tabular}{ll}
\toprule
Structure & Ingredients \\
\midrule
semigroup & set, operation, associativity \\
monoid & semigroup, identity element \\
group & monoid, inverses \\
abelian group & group, commutativity \\
ring & abelian group, monoid, distributivity \\
integral domain & ring, commutativity, no zero divisors \\
division ring & ring, inverses \\
field & integral domain, inverses \\
module & abelian group, ring, scalar action \\
vector space & module, field \\
graph & set, edge relation \\
tree & graph, connectedness, acyclicity \\
topological space & set, open sets \\
metric space & set, distance, triangle inequality \\
\bottomrule
\end{tabular}
\end{table}

\begin{table}[h]
\centering
\small
\caption{The 21 pairs of structures of the experiment on added requirements
(\cref{sec:composition}), grouped by the requirement they add.}
\label{tab:steps}
\setlength{\tabcolsep}{3pt}
\begin{tabular}{p{1.8cm}p{6.0cm}}
\toprule
Requirement & Pairs \\
\midrule
inverses & monoid $\to$ group, commutative monoid $\to$ abelian group,
semiring $\to$ ring, ring $\to$ division ring, commutative ring $\to$
field, integral domain $\to$ field \\
commutativity & monoid $\to$ commutative monoid, group $\to$ abelian
group, ring $\to$ commutative ring, division ring $\to$ field \\
compactness & metric space $\to$ compact metric space, topological
space $\to$ compact space, Hausdorff space $\to$ compact Hausdorff space \\
completeness & metric space $\to$ complete metric space, normed space
$\to$ Banach space, inner product space $\to$ Hilbert space \\
finiteness & group $\to$ finite group, field $\to$ finite field,
measure space $\to$ finite measure space \\
Hausdorff separation & topological space $\to$ Hausdorff space, compact
space $\to$ compact Hausdorff space \\
\bottomrule
\end{tabular}
\end{table}

\section{Membership}
\label{app:asymmetry}

This appendix details the experiments of \cref{sec:aboutness}, which
ask whether the model, and concept vectors, represent whether an
object $x$ is an instance of a structure $c$. A \emph{near miss} is a
pair where $x$ is an instance of a parent of $c$ without being an
instance of $c$, such as the symmetric group $S_3$ for abelian group.

\subsection{Families, texts and minimal contrasts}
\label{app:texts}

{We group the structures we use into three
\emph{families}: algebra (semigroup, monoid, group, abelian group,
ring, integral domain, division ring, field), graphs (graph, tree,
forest, bipartite graph, planar graph) and topology (topological space,
metric space, compact metric space, manifold). Each of the 18 instances
of the bank ($\mathbb{Z}$, $\mathbb{Q}$, $S_3$, $K_5$, the circle,
\dots) is assigned to one family. We use three kinds of texts.
\renewcommand{\labelenumi}{(\roman{enumi})}
\begin{enumerate}
\item \emph{Instance texts}: the corpora of the 18 instances
(\cref{sec:method}), which never name a structure of the instance's
family, keeping 76--90 texts each. For an instance $x$ and a
structure $c$, we average the scores $s_c(t)$ of the texts $t$ of $x$,
and standardize these averages over the 18 instances, so that a
structure whose vector gives high scores to all texts does not weigh
more than the others. The \emph{membership AUC} is the AUC of these
standardized averages between the pairs $(x, c)$ where $x$ is an
instance of $c$ and the pairs where it is not. It is computed in the
same way with the word-statistics scores.
\item \emph{Statements}: 2{,}604 templated statements
``$x$ is (not) a $c$'', over the 18 instances, the structures of their
family, the structures of one other family (graphs for algebra and
topology, algebra for graphs) and 6 phrasings, read at the last token. Each phrasing has a
\emph{positive} form, which says that $x$ is a $c$, and a
\emph{negative} form, which says that $x$ is not a $c$. The most meaningful statements are the \emph{near misses}, that is
the statements in which $x$ is not an instance of $c$ but is an
instance of a parent of $c$. In a near miss, $x$ is an instance of a
concept close to $c$ in the inclusion graph, so that this closeness
alone cannot tell a near miss from a true statement. There are 264
near misses among the 2{,}604 statements.
\item \emph{Questions}: the
118 yes/no questions ``Is $x$ a $c$? Answer yes or no.'', for the 18
instances and the structures of their family. We read the questions at the last token of the name
of $x$, which we call the \emph{object token}. It comes before $c$ is
named, so that what is read there is what the model holds about $x$
before being asked about $c$, as attributes of a subject are
retrieved at its last token \citep{meng2022rome,geva2023dissecting}.
\end{enumerate}}

We call \emph{minimal contrast} a structure $c$ together with two
instances of a parent of $c$, only one of which is an instance of $c$.
We use six of them: abelian group with $\mathbb{Z}$ and $S_3$, field
with $\mathbb{Q}$ and $\mathbb{Z}$, field with $\mathbb{C}$ and the
quaternions $\mathbb{H}$, integral domain with $\mathbb{Z}/5$ and
$\mathbb{Z}/6$, group with $\mathbb{Z}$ and $\mathbb{N}$, and tree
with the path $P_5$ and $K_5$. Each minimal contrast compares what is
read about the two instances, on instance texts or on statements, as
specified where it is used.

\subsection{Probe}
\label{app:probe}

\paragraph{Method.} \textit{(i) Probe.} We train a
\emph{probe}, meaning a linear classifier (via logistic regression)
that predicts, from the activation of the last token of a statement
(\cref{sec:method}), whether this statement is true. For each structure
$c$, the probe is trained on the statements about the other structures,
and applied to the statements about $c$. A statement ``$x$ is a $c$''
is true exactly when $x$ is an instance of $c$. A probe trained on
statements about $c$ could learn itself which objects are instances of
$c$. A probe that has never seen $c$ can only succeed if the model
represents being an instance of a concept in the same way for all the
structures, that is, as a relation between an object and a concept. We
do the same holding out one instance at a time. We evaluate the probe
by the AUC of its outputs between true and false statements, computed
only on the near misses and on the statements in which $x$ is an
instance of $c$, in both forms: since a near miss is an instance of a parent
of $c$, a probe that separates them distinguishes $c$ from its parent,
and so reads the asymmetry of this relation. The same probe is trained
on word counts.

\textit{(ii) Minimal contrasts.} We apply the probe, with the
structure held out, to the statements of the six minimal contrasts.
For a contrast between an instance $x$ of $c$ and an instance $x'$ of a
parent of $c$ that is not an instance of $c$, we take, in each of the 6
phrasings and each of the 2 forms, the statement about $x$ and the
statement about $x'$, such as ``$x$ is a $c$'' and ``$x'$ is a $c$''
in the positive form, and compare the probabilities of being true that
the probe gives them. This makes 12 comparisons per contrast. A
comparison is right when the true statement has the higher probability,
that is, the statement about $x$ in the positive form, and the
statement about $x'$ in the negative form.

\textit{(iii) Knowledge about the object alone.} The probe could
also succeed by reading only the object, and not the pair of the object
and the structure. Some objects are instances of many structures and
others of few, so that a probe reading only how many structures an
object belongs to would often be right. We exclude this in two ways.
First, we compute the AUC of the probe within the statements about one
object, between the true and false ones, separately for the positive
and the negative statements, with the probe trained on the other
objects. What depends on the object alone is then the same for all
these statements, so that knowledge about the object alone cannot
separate them. Second, we replace the activation of each statement by
its mean over the statements about the same object, with the same
phrasing and the same form, positive or negative. This removes
information about the structure while keeping information about the
object, so that a probe trained on these activations can only read the
object.

\paragraph{Results.} \textit{(i) Probe.} The AUC of the probe is about 0.85 on R1-Qwen-7B (layers 20 and
24) and 0.98 on R1-Qwen-32B (layers 44 and 56), and varies by less
than 0.01 between these layers, and between holding out one structure and
holding out one instance. The same probe on word
counts reaches 0.58 with one instance held out, and 0.20 with one
structure held out.

\textit{(ii) Minimal contrasts.} With one structure held out, the
probe gets 66 and 68 of the 72 comparisons right on R1-Qwen-7B (layers
20 and 24), and all 72 on R1-Qwen-32B from layer 28 on.

\textit{(iii) Knowledge about the object alone.} The AUC of the
probe within the statements about one object is 0.90--0.91 on
R1-Qwen-7B and 0.98 on R1-Qwen-32B, at the same layers as in
\textit{(i)}, against 0.72 for word statistics. The probe
trained on the activations without the structure has an AUC of
0.41--0.43, below chance. Since the minimal
contrasts fix the structure and the statements about one object fix the
object, neither alone explains the probe, which therefore reads the
pair of the object and the structure.

\subsection{Scoring objects on concept vectors}
\label{app:scoring}

For a structure $c$, we score the
objects of its family on the concept vector of $c$ in three ways. (a)
Each instance of the bank is scored by the mean score on $c$ of its
texts. (b) Each of the 6 positive statements ``$x$ is a $c$'', one per
phrasing, is scored separately. (c) Each object is scored by the score
on $c$ of its object token in the questions, which we ask about 77
objects: the 18 instances of the bank, and 59 objects of the three
families that are not in the bank, and therefore are not associated
with texts or statements. In each case, we take the AUC between the
scores of the instances of $c$ and those of the other objects of the
family, and we average it over the structures. We also use the six minimal
contrasts. A minimal contrast is right on the instance texts when the
mean score on $c$ of the texts of the instance of $c$ is higher than
that of the texts of the other instance, with a 95\% bootstrap
confidence interval over the texts that excludes zero. We do the same with
the word-statistics concept vector.

\subsection{Topic or property?}
\label{sec:topic}

\paragraph{Question.} Does a text about $S_3$ score lower on
\emph{abelian group} than a text about $\mathbb{Z}$?

\paragraph{Operationalization.} We score the instance texts and
the statements (\cref{sec:aboutness}) on the concept vectors of the
structures.

\paragraph{Results.} On instance texts, the membership AUC is
0.66--0.67, below word statistics (0.72). The minimal contrasts fail on
both models. Four show no effect, and the contrast between $S_3$ and
$\mathbb{Z}$ goes significantly the wrong way, with $S_3$ texts scoring
higher on abelian group. Only $P_5$ against $K_5$ on tree works. On
statements, the structure's vector does not separate members from near
misses (AUC 0.50--0.52 on R1-Qwen-7B and 0.47--0.55 on R1-Qwen-32B, at every
layer), and it ignores negation. ``$\mathbb{Z}$ is not a field'' scores
like ``$\mathbb{Z}$ is a field'' (the difference between positive and
negative statements is 0.00 to $+0.09$ standard deviations on R1-Qwen-7B,
and within $\pm 0.17$ on R1-Qwen-32B). $S_3$ scores above $\mathbb{Z}$ on
abelian group in every phrasing.

\paragraph{Interpretation.} Concept vectors measure aboutness, which
includes what a text evokes without naming it. $S_3$
is the standard first example of a non-commutative group, so texts
about it are about commutativity, and they score accordingly.
Mathematics is a domain where aboutness and truth come apart. In
domains such as emotions or personas, a text about anger is usually
also an angry text, and the two cannot be told apart.

\section{Concept vectors and writing vectors along chains of thought}
\label{app:chains}

This appendix studies the two kinds of vectors of the paper on the
model's own reasoning: whether concept vectors track the techniques
that a chain of thought uses, and whether writing vectors read the
concepts that a chain is about to name.

\subsection{Do concept vectors track the model's own reasoning?}
\label{sec:cot}
\label{app:cot}

\paragraph{Question.} If concept vectors measure aboutness, they should
still say which technique a chain of thought is about, for instance
induction, even when the chain never writes the word. Do they, and do
they say more than the words of the chain?

\paragraph{Operationalization.} Both models solve 80 competition
problems labelled with the bank's techniques \citep{yang2026beyond} and
200 MATH problems labelled by subject \citep{hendrycks2021measuring},
for 360 chains of thought of up to 8{,}192 tokens. Technique labels come
from rules and, separately, from a judge that tags which techniques
each chain actually uses. At every token we project the activation on
all concept vectors. The detection AUC compares a technique's mean
activation on chains that use it with chains that do not, pooled and
within the problem's domain, against a null obtained by shuffling the
labels. To measure timing, we align activation to the technique's
first mention. A regression asks whether activation predicts a correct
answer, controlling for difficulty and length. The word-statistics concept vector is
projected on the chain's text, and the model's scores are residualized
on it.

\paragraph{Results.} Techniques are detected, with AUC 0.74--0.80
pooled and 0.63--0.72 within domain (null 0.53--0.58), on both models
and both label sets. Techniques that the chain never names are detected
at 0.67--0.75, and MATH subjects at 0.82--0.94 except algebra.
Activation rises at and after the first mention ($+0.4$ to $+0.8$
standard deviations) and not before it, and it does not predict
correctness ($+0.04 \pm 0.14$ on R1-Qwen-7B, $+0.00 \pm 0.14$ on R1-Qwen-32B).
The words do as well. Word statistics detect techniques at 0.78--0.80
pooled and 0.66--0.71 within domain, and techniques that are never
named at 0.70--0.73, because an unnamed induction still leaves ``base
case'' and ``assume it holds for $n$'' in the text. Residualized on the
words, the model is at chance within domain (0.41--0.50) on both models
and both label sets. \Cref{tab:cot} gives the full table.

\paragraph{Interpretation.} Along its own reasoning, the model's state
tracks what the chain is about, as the words do. These measures see
nothing in it beyond the words. It does not anticipate the next
technique, and it does not predict success.

\Cref{tab:cot} gives the detection AUC for each model
and label set. Techniques are labelled by rules on the problem and,
separately, by the judge, which tags the techniques each chain actually
uses. ``Within domain'' compares chains on problems of the same domain
only, which removes the detection of the domain itself.

\begin{table}[h]
\centering
\small
\caption{Technique detection along chains of thought: AUC pooled /
within domain, for word statistics, for the model's concept vectors,
and for the model after residualizing on word statistics (residual).}
\label{tab:cot}
\setlength{\tabcolsep}{3.5pt}
\begin{tabular}{lccc}
\toprule
 & Word statistics & Model & Residual \\
\midrule
\multicolumn{4}{l}{R1-Qwen-7B} \\
\quad rule labels & 0.80 / 0.66 & 0.74 / 0.62 & 0.50 / 0.42 \\
\quad judge labels & 0.78 / 0.69 & 0.79 / 0.71 & 0.56 / 0.50 \\
\multicolumn{4}{l}{R1-Qwen-32B} \\
\quad rule labels & 0.80 / 0.71 & 0.74 / 0.63 & 0.50 / 0.41 \\
\quad judge labels & 0.78 / 0.71 & 0.80 / 0.71 & 0.56 / 0.50 \\
\bottomrule
\end{tabular}
\end{table}

The label-shuffle null is 0.50--0.58 depending on the set. The small
pooled residual of 0.56 ($p = 0.001$) comes from differences between
domains and disappears within domain. Techniques the chain never names
are detected at 0.67--0.75 by the model and 0.70--0.73 by the words, and
naming adds about 0.09. MATH subjects are detected at 0.82--0.94 by the
model except algebra (0.59--0.63), and at 0.71--0.95 by the words.
Activation rises at and after the first mention ($+0.4$ to $+0.8$ standard deviations)
and not before it, and it does not predict a correct answer when
difficulty and length are controlled ($+0.04 \pm 0.14$ on R1-Qwen-7B,
$+0.00 \pm 0.14$ on R1-Qwen-32B). R1-Qwen-32B solves 30\% of the competition
problems and 82\% of the MATH problems.

\subsection{Do writing vectors read?}
\label{sec:writeread}

\paragraph{Question.} Concept vectors read a concept and do not write
it. Does the converse hold? The writing vector of the word
``induction'' makes the model write it. Does it also tell, along a
chain of thought, that the model is about to use induction?

\paragraph{Operationalization.} We use the chains of thought of R1-Qwen-7B
from \cref{sec:cot} and take the first mention of each concept that the
problem statement does not name (856 first mentions of 69 concepts).
From 1 to 128 tokens before each first mention, we project the
activation at layer 16 on the writing vector of the concept's name, and
compute the AUC against two controls: other chains at a similar
position (across chains), and the same chain more than 256 tokens
earlier (within a chain). We compare with the logit lens, which reads
a layer directly through the output unembedding, and with word statistics,
the TF-IDF vector of the preceding 256 tokens. To find out whether what is
read is used, we cut 240 chains 16 or 64 tokens before a first mention,
sample four continuations of each, and remove the writing vector at
layers 8--24, either from the positions before the cut or from the
newly written tokens. The control removes another concept's writing
vector.

\paragraph{Results.} Across chains, the words read the upcoming concept
better than the writing vector (AUC 0.68 against 0.64, 64 tokens
ahead), because different problems are about different things. Within
a chain, where the topic is fixed, the writing vector reads the concept
64 tokens ahead at 0.59 [0.57, 0.61], against 0.55 for the logit lens
and 0.53 for the words. The signal falls to 0.55 at 128 tokens and
disappears two to three paragraphs ahead, and residualizing on the
words changes it by at most 0.03. Removing the writing vector from the
newly written tokens lowers the rate at which the model writes the
word by 0.33 to 0.41 relative to the control. Removing it only from the
positions before the cut lowers this rate by 0.07 [0.02, 0.12] with a
cut 16 tokens ahead, and by 0.01 [$-0.04$, 0.06] with a cut 64 tokens
ahead. The continuations stay coherent and express the idea without the
word.

\paragraph{Interpretation.} The writing vector of a concept reads it,
but weakly and locally. It reads the approach of the word, about a
paragraph ahead and within a chain whose topic is already fixed, and
what it reads there is barely used to produce the word. The two kinds
of vector are thus complementary. A concept vector reads what a text is
about and does not write it, and a writing vector writes the word and
reads little beyond its imminent use.

\end{document}